\documentclass[conference]{IEEEtran}

\usepackage{graphicx}
\usepackage{url}
\usepackage{algorithm}
\usepackage{algorithmic}
\usepackage{subfigure}
\usepackage{cite}
\usepackage{amsmath}
\usepackage{amssymb}

\newtheorem{theorem}{Theorem}
\newtheorem{lemma}{Lemma}
\newtheorem{definition}[theorem]{Definition}

\usepackage{booktabs}

\begin{document}

%

\title{Transfer Learning across Networks for Collective Classification}


\author{%
{Meng Fang{\small $^{\dag}$}, Jie Yin{\small $^{\ddag}$}, Xingquan Zhu{\small $^{\S}$} }%
\vspace{1mm}\\
\fontsize{10}{10}\selectfont\itshape
$~^{\dag}$ Centre for Quantum Computation \& Intelligent Systems, FEIT,
University of Technology, Sydney, Australia\\
\fontsize{9}{9}\selectfont\ttfamily\upshape
\fontsize{10}{10}\selectfont\rmfamily\itshape
$~^{\ddag}$Computational Informatics, CSIRO, Australia\\
$~^{\S}$Dept. of Computer \& Electrical Engineering and Computer Science, Florida Atlantic University, USA\\

\fontsize{10}{10}\selectfont\ttfamily\upshape
Meng.Fang@student.uts.edu.au; Jie.Yin@csiro.au; xzhu3@fau.edu
}

\maketitle

\begin{abstract}
This paper addresses the problem of transferring useful knowledge from a source network to predict node labels in a newly formed target network. While existing transfer learning research has primarily focused on vector-based data, in which the instances are assumed to be independent and identically distributed, how to effectively transfer knowledge across different information networks has not been well studied, mainly because networks may have their distinct node features and link relationships between nodes. In this paper, we propose a new transfer learning algorithm that attempts to transfer \emph{common latent structure features} across the source and target networks. The proposed algorithm discovers these latent features by constructing label propagation matrices in the source and target networks, and mapping them into a shared latent feature space. The latent features capture common structure patterns shared by two networks, and serve as domain-independent features to be transferred between networks. Together with domain-dependent node features, we thereafter propose an iterative classification algorithm that leverages label correlations to predict node labels in the target network. Experiments on real-world networks demonstrate that our proposed algorithm can successfully achieve knowledge transfer between networks to help improve the accuracy of classifying nodes in the target network.

\end{abstract}

\begin{IEEEkeywords}
Network; Transfer learning;

\end{IEEEkeywords}

%
\IEEEpeerreviewmaketitle

With recent advance in Web 2.0 technology, information networks, such as social networks, communication networks and bibliographic networks, are becoming ubiquitous in our daily life. Examples include the friendship network in \emph{Facebook}, co-author networks in \emph{DBLP} and citation networks in \emph{PubMed} for biomedical articles. Such networks have common properties that they all contain different kinds of entities which interact with one another. Accordingly, an information network is represented as a large graph, in which nodes denote entities or instances (\textit{e.g.}, users or scientific publications) and links denote relationships between nodes (\textit{e.g.}, friendship or citation relationships). To analyze such networks, an important task is to predict the labels of nodes in the networked data, which is commonly achieved by exploiting label correlations through collective classification \cite{neville2000iterative,sen2008collective}.

Despite the abundance of networked data, labels are usually very expensive and time consuming to obtain, particularly for newly formed information networks or any new/emerging disciplines in an existing network. In the meanwhile, it is not uncommon that plenty of labeled data exists in some different but related domains. To address this situation, transfer learning has emerged as a new machine learning framework that explores knowledge from auxiliary source domains to facilitate a new learning task in the domains of interest \cite{pan2010transfer}. The basic idea behind transfer learning is that the involved domains share some common latent factors, which can be uncovered and exploited using different techniques as the bridge for knowledge transfer. Most of the existing works on transfer learning have mainly considered traditional vector-based data \cite{Dai2007Boosting,Gao2008Knowledge,Pan2010Crossdomain}, in which each instance is represented by a multi-dimensional feature vector and the instances are assumed to be independent and identically distributed (i.i.d.). However, little research work has been done to address the problem of effective and reliable knowledge transfer across different networks.

Performing transfer learning across different networks poses a number of new challenges, due to the characteristics of networked data. First, the source and target networks can be heterogeneous in nature, because they are formed by different reasons and driven by different applications and user groups. The two networks can be distinct in that their nodes represent different entities, and the associated links indicate different relationships between nodes. For example, a \emph{Facebook} network indicates friendship relationship between users and a \emph{PubMed} network represents the citation relationships between scientific publications. A friendship is clearly different from a citation relationship in the sense that the former relies on social interactions between users, whereas the latter is more focused on the content sharing between scientific publications. Even in the case that links may share similar relationships across networks, each network may reveal different features for its own nodes. For two citation networks, \emph{CiteSeer} and \emph{PubMed}, the former mainly comprises academic papers in computer science while the latter focuses on biomedical articles. Clearly, the feature spaces of the nodes from the two networks, \textit{i.e.}, keywords in paper titles, can be largely different with limited overlap. Thus, the knowledge on node features is not necessarily transferable across different networks, and in the presence of network heterogeneity, discovering common latent factors using the overlap of node features would render sub-optimal results, as traditional transfer learning does. Second, in the context of networked data, instances are not independent, but are connected by links between each other to form a network. As a result, the labels of connected nodes are correlated in a local neighborhood. This indicates that, closely connected nodes tend to share the same label and nodes on the same substructure are likely to share the same label. Such correlations should be preserved when common latent factors are discovered as the bridge for knowledge transfer across networks. Based on these observations, one key research question is, what information can be transferred from the source network to build an accurate classifier in the target network?

In this paper, we propose a novel approach to address the problem of transfer learning across information networks. Our key idea is to discover and transfer some common structure knowledge from the source network to the target network. Specifically, we construct a label propagation matrix that captures the influence of structure information on the labels of connected nodes in a network. Based on this, we design an optimization problem to uncover \emph{latent structure features} which can capture common structure patterns shared by the source and target networks. These latent features are domain-independent, and can thus serve as generic features transferred from the source network to boost the classification task in the target network. With domain-independent, latent structure features and domain-dependent node features, we develop an iterative classification algorithm (ICA) that makes use of label correlations to predict the labels of nodes in the target network. We have conducted extensive experiments on four real-world networks and demonstrated that our proposed transfer learning algorithm can significantly improve the accuracy of classifying nodes in the target network.

\section{Related Work}
In this section, we briefly review related studies on collective classification over networked data and existing research works on transfer learning.

\textbf{Collective Classification:}
Collective classification has recently attracted significant attention for classifying relational data in the data mining area \cite{sen2008collective,CautiousCC2009McDowell}. Networked data is one typical type of relational data, in which instances are represented as nodes and the relationships between nodes are represented as edges. Collective classification exploits dependencies between instances, which makes it one of the most favorable classification methods for networked data.

Approaches to collective classification can be roughly grouped into global methods and iterative methods. Global methods aim to train a classifier that seeks to optimize a global objective function, often based on a Markov random field. These methods are usually computationally expensive, which limits their applicability to large-scale, real-world networked data. On the other hand, iterative methods employ an iterative process whereby a local classifier predicts labels for each node by using node features and relational features derived from the current label predictions. After that, a collective inference algorithm recomputes the class labels, which will be used in the next iteration.

Iterative classification algorithm (ICA) is an iterative method that is widely applied and extended in many studies ~\cite{neville2000iterative,bilgic2007combining,bilgic2010active}. The basic assumption of ICA is that, given the labels of a node's neighbors, the label of the node is independent of the features of its neighbors and non-neighbors, and the labels of all non-neighbors. In ICA, each node is expressed by combining the node features and relational features constructed by using the labels of all the neighbors of the node. The relational features can be computed by using an aggregation function over the neighbors, such as $\mathtt{count}$, $\mathtt{mode}$, $\mathtt{proportion}$ and so on. Based on the node features and relational features, ICA trains a classifier and iteratively updates the predictions of all nodes by using the predictions for node with unknown labels. This process continues until the algorithm converges. In this work, we adopt an ICA-like algorithm to perform collective classification with focuses on transferring structure knowledge from the source network to improve collective classification accuracy on the target network, under the assumption that the number of labeled nodes is very limited.

\textbf{Transfer Learning:}
Transfer learning has emerged as a new machine learning paradigm that exploits labeled data in the source domain to build an accurate classifier in the target domain, where the labeled data in the target domain is very limited \cite{pan2010transfer}. According to the type of information to be transferred, transfer learning approaches can be broadly summarized into three categories. The first category is based on instance transfer \cite{Dai2007Boosting,Dai2007transferring}, in which certain parts of the instances in the source domain can be reused for learning in the target domain via instance weighting. TrAdaBoost \cite{Dai2007Boosting} is one typical example of such methods. TrAdaBoost adjusts the contributions of training instances by giving larger weights to the instances from the source domain that are more similar to the target instances. These methods usually require that different domains share the same feature space and label space, so that the same classifier can be trained on both domains to perform classification. The second category is the parameter transfer approach \cite{Gao2008Knowledge}, which assumes that the source and target learning tasks share similar parameters or prior distributions of the models, and thus transferring these parameters or priors can help improve the learning task in the target domain. The third family of methods aim to learn a good latent feature representation shared by two domains \cite{Blitzer2006Domain,Pan2010Crossdomain}, where the knowledge used to transfer across domains is encoded into the learned feature representation.

While a large amount of research has been proposed for transfer learning, existing studies have focused on conventional vector-based data, in which each instance is represented by a multi-dimensional features vector, and all instances are assumed to be independent and identically distributed (i.i.d.). Recently, some early attempts intend to deal with transfer learning in relational domains, where the instances are non-i.i.d. and can be represented by multiple relations. Mihalkova et al. \cite{Mihalkova2007Mapping} proposed a TAMAR algorithm to transfer relational knowledge with Markov Logic Networks (MLNs) across relational domains. In MLNs, entities are represented by predicates and their relationships are represented in first-order logic. TAMAR tries to map an MLN learned for a source domain to the target domain based on weighted pseudo log-likelihood measure, and the mapped structure is further revised as a relational model for inference in the target domain. Another work \cite{ye2013predicting} proposed an approach to leveraging the edge sign information across signed social networks for edge sign prediction.


To the best of our knowledge, our work is the first research endeavor focusing on transferring knowledge across information networks to predict node labels, where the feature space of the nodes and the node labels of two networks can be largely different. Our proposed method falls into the third category of transfer learning approaches, which attempts to discover common latent structure features shared by the source and target networks. Being domain-independent, these latent features are considered as the bridge to transfer knowledge across different networks.

\section{Problem Definition}
We focus on an inductive transfer learning setting, where the nodes in the source network are fully labeled, while the target network only has a small number of labeled nodes. We consider one source network $G_s$ and one target network $G_t$ for our classification task. The target network is represented as a graph $G_t=(\mathcal{V}^{u}_t, \mathcal{V}^{l}_t, \mathcal{E}_t)$, where $\mathcal{V}^{l}_t$ denotes the small set of labeled nodes in the network and $\mathcal{V}^{u}_t$ denotes the set of nodes whose class labels are unknown and need to be predicted. $\mathcal{E}_t$ denotes the set of edges connecting the nodes. Each node $v^{i}_{t} \in \mathcal{V}^{u}_t \cup \mathcal{V}^{l}_t$ is described by a feature vector $\mathbf{x}^{i}_t$. For a node $v^{i}_{t} \in \mathcal{V}^{l}_t$, it is also associated with a class label $y^{i}_{t} \in \mathcal{Y}_t$, where $\mathcal{Y}_t$ denotes a set of class labels in the target domain.

In transfer learning setting, we also have a fully labeled source network which is represented as $G_s=(\mathcal{V}^{l}_{s},\mathcal{E}_s)$, where $\mathcal{V}^{l}_{s}$ denotes the set of labeled nodes and $\mathcal{E}_s$ denotes the set of edges between the labeled nodes. Each node $v^{i}_{s} \in \mathcal{V}^{l}_{s}$ is associated with a feature vector $\mathbf{x}^{i}_s$ and a class label $y^{i}_{s} \in \mathcal{Y}_s$, where $\mathcal{Y}_s$ denotes a set of class labels in the source domain. Note that, in our transfer learning problem, we do not require nodes in $G_s$ and $G_t$ to share the same feature space and label space.


Given the source network $G_s=(\mathcal{V}^{l}_{s},\mathcal{E}_s)$ and the target network $G_t=(\mathcal{V}^{u}_t, \mathcal{V}^{l}_t, \mathcal{E}_t)$, the goal of our transfer learning task is to (1) uncover common latent factors shared by the source and target networks, and (2) leverage these latent factors to help predict unlabeled nodes $v^{i}_{t} \in \mathcal{V}^{u}_t$ in the target network with one of class labels $y^{i}_{t} \in \mathcal{Y}_t$.

\section{The proposed Algorithm}
The most important issue of our transfer learning problem is to identify knowledge/patterns which are transferable across different networks. Unlike traditional transfer learning problems, networks can contain nodes with different content features, and the label space of the networks can be totally different. For networked data, nodes are connected by links to form a network, closely connected nodes tend to have the same label, and nodes sharing the same structure patterns are likely to have the same label. Therefore, we propose to transfer structure information from the source network to the target network for predicting node labels in the target network.

Our proposed algorithm consists of two major parts: learning latent structure features and carrying out collective classification. In order to learn latent structure features, we first define a label propagation matrix which reveals the influence of structure information on the labels of the nodes that are connected with each other. Based on this, we formulate and solve an optimization problem for discovering common latent structure features. These latent features serve as domain-independent features that capture the common structure patterns shared by networks. Together with domain-dependent node features, we further develop a transfer learning algorithm for collective classification.

In the following, we first define the label propagation matrix and propose an objective solution to learn the latent structure features. Then we detail our proposed transfer learning algorithm for collective classification.

\subsection{Label Propagation Matrix}
Our work is to find ``good'' feature representations shared by different networks to minimize domain divergence and classification errors. Although nodes in different networks can have different feature space and label space, they do share some common structure patterns, based on which nodes can have the same label. To capture such information, we propose to construct a label propagation matrix to model how network structures influence the labels of connected nodes in the network.

Specifically, we borrow the idea from semi-supervised learning. Semi-supervised learning~\cite{zhu2003semi,chapellelabel} builds a graph in which nodes represent data points and edges represent similarities between points. We use the geometry of network to represent the similarities between nodes. Those similarities are given by a weighted matrix $W$, where $W_{ij}$ is non-zero if $\mathbf{x}_i$ and $\mathbf{x}_j$ are neighbors in the network. Thus we have
\begin{align}
 W_{ij} = \left\{
  \begin{array}{l l}
    1 & \quad \text{$\mathbf{x}_i$ and $\mathbf{x}_j$ are neighbors,}\\
    0 & \quad \text{otherwise.}
  \end{array} \right.
  \label{eq-w}
\end{align}
An alternative weight matrix can be given by a Gaussian kernel with width $\sigma$:
\begin{equation}
W_{ij}=\exp{\left\{-\frac{\|x_i-x_j\|}{2\sigma^2}\right\}}
\label{eq-w-gaussian}
\end{equation}
where $W_{ij}$ is symmetric positive matrix given by a symmetric positive function $W_X$.

Given a graph $G$, we consider a process of propagating the labels on the graph, for both labeled nodes 1, 2, $\dotsc$, $l$, and unlabeled nodes $l+1$, $\dotsc$, $n$. Each node propagates its label to its neighbors, and the propagation process is repeated until reaching to convergence.

Based on this process, we introduce a new matrix, named label propagation matrix, for expressing the propagated correlations between connected nodes in a network, inspired by the idea of semi-supervised learning~\cite{zhu2002learning,zhou2004learning}. We assume that a node $i$ receives a contribution from its neighbors $\mathcal{N}_i$, and also retains an additional contribution given by its initial value. The process is given in Algorithm \ref{alg-propagation} below.



\begin{algorithm}
\caption{Label Propagation Process}
\label{alg-propagation}
\begin{algorithmic}[1]
\STATE Calculate the affinity matrix $W$ by using Eq.(\ref{eq-w-gaussian}) if $i \leq j$ and $W_{ii}=0$

\STATE Calculate the diagonal degree matrix of $D$: $D_{ii}=\sum_{j}W_{ij}$


\STATE Calculate the matrix $\mathcal{L}=D^{-1/2}WD^{-1/2}$

\STATE Give a parameter $\alpha \in [0,1)$

\WHILE{ $\hat{Y}$ is not convergence}

\STATE $\hat{Y}^{(t+1)}=\alpha\mathcal{L}\hat{Y}^{(t)}+(1-\alpha)\hat{Y}^{(0)}$

\ENDWHILE
\end{algorithmic}
\end{algorithm}

We now prove the convergence of Algorithm~\ref{alg-propagation}.

\begin{proof}
From Algorithm~\ref{alg-propagation}, the iteration equation is
\begin{equation}
\hat{Y}^{(t+1)}=\alpha\mathcal{L}\hat{Y}^{(t)}+(1-\alpha)\hat{Y}^{(0)},
\label{eq-pm-iteration}
\end{equation}
then we have
\begin{equation}
\hat{Y}^{(t+1)}=(\alpha\mathcal{L})^{t}\hat{Y}^{(t)}+(1-\alpha)\sum_{i=0}^{t}(\alpha\mathcal{L})\hat{Y}^{(0)}.
\end{equation}
The Laplacian matrix $\mathcal{L}$ is similar to $S=D^{-1}W=D^{-1/2}\mathcal{L}D^{1/2}$
and they have the same eigenvalues. Since $S$ is a stochastic matrix, its
eigenvalues are within the range of $[-1,1]$. Given that $0 < \alpha < 1$, we have
\begin{equation}
\lim_{t \to \infty }(\alpha\mathcal{L})^{t}=0,
\end{equation}
and
\begin{equation}
\lim_{t\to \infty }\sum_{i=0}^{t}(\alpha\mathcal{L})^{i} = (I-\alpha\mathcal{L})^{-1}.
\end{equation}
So when $t \to \infty$ we have
\begin{equation}
\hat{Y}^{(t)}=\hat{Y}^{\infty}=(1-\alpha)(I-\alpha\mathcal{L})^{-1}\hat{Y}^{0}.
\label{eq-pm-proof-inf}
\end{equation}
Now we can see there exists the convergence when $t \to \infty$ and the convergence rate depends on specific properties of the graph, that is, the eigenvalues of the Laplacian matrix.
\end{proof}

The main part of Algorithm~\ref{alg-propagation} is the iteration process (as defined by Eq.~(\ref{eq-pm-iteration})). The first term of Eq.~(\ref{eq-pm-iteration}) indicates that each data point receives the information from its neighbors. The second term of Eq.~(\ref{eq-pm-iteration}) indicates that the data point is also influenced
by its initial label information. Now we focus on how $\hat{Y}^{(t)}$ is influenced and becomes stable when nodes receive information from the neighbors and their initial labeling information. The Proof above indicates that we can compute $\lim_{t \to \infty}\hat{Y}^{(t)}$ directly without doing iterations using Eq.~(\ref{eq-pm-proof-inf}).
Accordingly, we define the label propagation matrix as follows:
\begin{equation}
\hat{Y}^{(t)}=\mathcal{P}\hat{Y}^{(0)},
\end{equation}
where $\mathcal{P}=(I-\alpha\mathcal{L})^{-1}$. Here, $\mathcal{P}$ is the label propagation matrix and it translates $\hat{Y}^{(0)}$ to its convergence status $\hat{Y}^{\infty}$. $\mathcal{P}$ is a nonnegative matrix.

We give a simple proof to show that $\mathcal{P}$ is a nonnegative matrix. We let $Q=I-\alpha\mathcal{L}$, and thus $\mathcal{P}=Q^{-1}$. Because $0<\alpha<1$, we have $Q_{ii}=1$ and $\sum_{j\neq i}Q_{ij} < -1 $. We can translate $[Q~I]$ to $[I~Q^{-1}]$ by using
elementary row operations. Because only pivot elements are 1 and others are negative $(-1,0)$, we only need to do row addition and the elements which are not pivot elements can be zero. As pivot elements are in $(0,1)$, we obtain row multiplication by multiplying a positive value for each pivot element. Therefore the left parts of elementary raw operations on $[Q~I]$ are always positive, \textit{i.e.} $\mathcal{P}$ is a nonnegative matrix.

\subsection{Learning latent structure features}
\label{subsec-latent-features}

Given the source network $G_s=(\mathcal{V}^{l}_s,\mathcal{E}_s)$, and the target network $G_t=(\mathcal{V}^{u}_t, \mathcal{V}^{l}_t, \mathcal{E}_t)$, we can calculate their label propagation matrices, respectively. Note that $G_s$ is fully labeled and we have $\mathcal{Y}_s$. We can compute the propagation matrix $\mathcal{P}_s$. For partially labeled target graph $G_t$, we can compute the propagation matrix $\mathcal{P}_t$.

Given $\mathcal{P}_s$ and $\mathcal{P}_t$, we propose to use nonnegative matrix factorization~\cite{seung2001algorithms} to construct latent propagation features through factorizing $\mathcal{P}_s$ and $\mathcal{P}_t$ under the same space. For $\mathcal{P}_s$, we have
\begin{equation}
\min \| \mathcal{P}_s -F_sR^{T}_s\|^2,
\label{eq-factorization-source}
\end{equation}
and for $\mathcal{P}_t$ we have
\begin{equation}
\min \| \mathcal{P}_t -F_tR^{T}_t\|^2.
\label{eq-factorization-target}
\end{equation}
However, the two factorizations below are very limited because $F_s$ and $F_t$, $R_s$ and $R_t$ have different scales and dimensions. As a result, it is very difficult to find shared latent feature space directly. Instead, we define $R_s$ with $R_sA^{T}$, and similarly, $R_t$ with $R_tA^{T}$. Therefore, we can rewrite Eq.(\ref{eq-factorization-source}) and  Eq.~(\ref{eq-factorization-target}) as
\begin{equation}
\min \| \mathcal{P}_s -F_sAR^{T}_s\|^2,
\label{eq-factorization-source-tri}
\end{equation}
and
\begin{equation}
\min \| \mathcal{P}_t -F_tAR^{T}_t\|^2.
\label{eq-factorization-target-tri}
\end{equation}
where the matrix $A$ is common latent features for both networks and ensures the extracted latent structure features can be represented by the same space.

To discover common latent features shared by networks, we define our optimization objective function as
\begin{equation}
\begin{aligned}
& \min & & \| \mathcal{P}_s -F_sAR^{T}_s\|^2+\| \mathcal{P}_t -F_tAR^{T}_t\|^2+\beta\|A\|^2, \\
& \text{s.t.} & & \sum_{j}F_{s(.j)}=1, \sum_{j}R_{s(.j)}=1, \\
& & & \sum_{j}F_{t(.j)}=1, \sum_{j}R_{t(.j)}=1, \\
& & & F_s, R_s \in \mathbb{R}^{M\times k}_{+}, F_t, R_t \in \mathbb{R}^{N\times k}_{+}, A \in \mathbb{R}^{k\times k}_{+}.
\end{aligned}
\label{eq-opt}
\end{equation}
In the above objective function, the first two terms are two matrix factorizations where $\mathcal{P}_s \approx F_sAR^{T}_s$ and $\mathcal{P}_t \approx F_tAR^{T}_t$. $A$ is latent structure features for both networks. $F_s$ and $F_t$ are two new feature representations in the latent space. $R^{T}_s$ and $R^{T}_t$ are two additional factors that absorb  different scales of $\mathcal{P}$, $F$ and $A$. $\|A\|^2$ is a penalty when $\|A\|$ is too large. $\beta$ balances the trade-off between the complexity of $A$ and two factorization terms.
Since all variables are nonnegative, a larger value of $A$ would make other variables $F_s$, $R_s$, $F_t$ and $R_t$ smaller. Especially, extremely large values in $A$ would make lots of elements in other variables be close to zeros. As a result, the new feature representation of nodes in the target network would have many missing values. Consequently, it would degrade the node classification accuracy. Therefore, it is necessary to control the values in $A$ by adding a regularization term.

\subsubsection{Solving optimization}
Given the optimization function, we write Eq.~(\ref{eq-opt}) as
\begin{align*}
J &=\| \mathcal{P}_s -F_sAR^{T}_s\|^2+\| \mathcal{P}_t -F_tAR^{T}_t\|^2+\beta\|A\|^2, \\
&=\text{Tr}(\mathcal{P}_s^{T}\mathcal{P}_s-2\mathcal{P}_s^{T}F_sAR^{T}_s+R_sA^{T}F^{T}_sF_sAR^{T}_s) \\
&+\text{Tr}(\mathcal{P}_t^{T}\mathcal{P}_t-2\mathcal{P}_t^{T}F_tAR^{T}_t+R_tA^{T}F^{T}_tF_tAR^{T}_t) \\
&+\beta\text{Tr}(A^{T}A).
\end{align*}
We iteratively compute the variables for above function by updating one variable and letting others be fixed.

\textbf{Update $A$:} Fixing $\mathcal{P}_s, F_s, R_s,\mathcal{P}_t,F_t,R_t$ and given the constraint
$A \in \mathbb{R}^{k\times k}_{+}$, we introduce the  Lagrangian multipliers $\lambda_{A}$, $\lambda_{A} \in \mathbb{R}^{k\times k}$ and minimize the Lagrangian function
\begin{equation}
L(A,\lambda_{A})= J-\text{Tr}(\lambda_{A}A).
\end{equation}
The gradient of $L(A,\lambda_{A})$ with respect to $A$ is
\begin{eqnarray}
\frac{\partial L}{\partial A}=-2F_s^{T}\mathcal{P}_sR_s+2F_sF_s^{T}AR_s^{T}R_s \nonumber \\
-2F_t^{T}\mathcal{P}_tR_t + 2F_tF_t^{T}AR_t^{T}R_t +2\beta A-\lambda_{A}.
\end{eqnarray}
Then from the KKT complementarity condition we have
\begin{equation}
\frac{\partial L(A,\lambda_{A})}{\partial A}=0,
\end{equation}
\begin{equation}
\lambda_{A}A=0,
\end{equation}
and we can rewrite above function as
\begin{eqnarray}
(-F_s^{T}\mathcal{P}_sR_s+F_sF_s^{T}AR_s^{T}R_s-F_t^{T}\mathcal{P}_tR_t \nonumber \\
+ F_tF_t^{T}AR_t^{T}R_t +\beta A)A=0.
\label{eq-kkt-condition}
\end{eqnarray}
We solve the above coupled equations by using auxiliary function approach~\cite{seung2001algorithms}.
According to~\cite{seung2001algorithms}, the auxiliary function is defined as
\begin{definition}
$Z(h,h')$ is an auxiliary function for $F(h)$ if the conditions $Z(h',h) \geq F(h)$ and $Z(h,h)=F(h)$ are satisfied.
\end{definition}

According to Eq.~(\ref{eq-opt}) and ignoring
the fixed variables, we can define objective function as
\begin{eqnarray}
J(A)=-2\text{Tr}(F_s^{T}\mathcal{P}_sR_s)-2\text{Tr}(F_t^{T}\mathcal{P}_tR_t) \nonumber \\
+\text{Tr}(F_sF_s^{T}AR_s^{T}R_sA^{T}) \nonumber \\
+\text{Tr}(F_tF_t^{T}AR_t^{T}R_tA^{T}) + \beta\text{Tr}(A^{T}A).
\label{eq-aux-h}
\end{eqnarray}
From Eq.~(\ref{eq-aux-h}) we define the following function
\begin{eqnarray}
Z(A,A')= -2\text{Tr}(F_s^{T}\mathcal{P}_sR_sA)-2\text{Tr}(F_t^{T}\mathcal{P}_tR_tA) \nonumber \\
+\sum_{i,j}\frac{(F_s^{T}F_sA'R_s^{T}R_s)_{(ij)}A^{2}_{(ij)}}{A'_{(ij)}} \nonumber \\
+ \sum_{i,j}\frac{(F_t^{T}F_tA'R_t^{T}R_t)_{(ij)}A^{2}_{(ij)}}{A'_{(ij)}}+\sum_{i,j}\frac{(\beta A')_{(ij)}A^{2}_{(ij)}}{A'_{(ij)}}.
\label{eq-auxiliary}
\end{eqnarray}
This function is an auxiliary function of $J(F_s)$. We will give proof later. Firstly we give a Lemma from~\cite{ding2006orthogonal}.
\begin{lemma}
For any matrices $C \in \mathbb{R}_{+}^{n \times n}$, $D \in \mathbb{R}_{+}^{k \times k}$,$H \in \mathbb{R}_{+}^{n \times k}$,$H' \in \mathbb{R}_{+}^{n \times k}$ and $C$,$D$ are symmetric, the following inequality holds
\begin{equation}
\sum_{i,j}\frac{(CH'D)_{ij}H^{2}_{ij}}{H'_{ij}} \geq \text{Tr}(H^{T}CHD),
\end{equation}
\end{lemma}
and then we show the proof of auxiliary function.
\begin{proof}
According to Lemma 1 and the third term in Eq.~(\ref{eq-auxiliary}), we let $C=F_s^{T}F_s$, $D=R_s^{T}R_s$,
$H'=A'$ and $H=A$. We have
\begin{eqnarray}
\text{Tr}(H^{T}CHD)=\text{Tr}(A^{T}F_s^{T}F_sAR_s^{T}R_s),
\end{eqnarray}
where $\text{Tr}(A^{T}F_s^{T}F_sAR_s^{T}R_s)=\text{Tr}(F_sF_s^{T}AR_s^{T}R_sA^{T})$. Then we can show that the third term in $Z(A,A^{'})$ is always bigger than the third one in $J(A)$. In the same way we
can show that the fourth and fifth terms in $Z(A,A^{'})$ are always bigger than the fourth and
fifth terms in $J(A)$ respectively.
And they have the same first term and second term. Thus $Z(A,A') \geq J(A)$.

We verify that $Z(A,A)=J(A)$. We rewrite the third term in Eq.~(\ref{eq-auxiliary})
by setting $A'=A$ as follows
\begin{eqnarray}
\sum_{i,j}\frac{(F_s^{T}F_sAR_s^{T}R_s)_{(ij)}A^{2}_{(ij)}}{A_{(ij)}}  \nonumber \\
=\sum_{i,j}\frac{(A^{T}F_s^{T}F_sAR_s^{T}R_s)_{(ij)}A^{2}_{(ij)}}{A^{T}A_{(ij)}}  \nonumber \\
=\sum_{i,j}{(A^{T}F_s^{T}F_sAR_s^{T}R_s)_{(ij)}} \nonumber \\
=\text{Tr}(A^{T}F_s^{T}F_sAR_s^{T}R_s),
\end{eqnarray}
where $\text{Tr}(A^{T}F_s^{T}F_sAR_s^{T}R_s)=\text{Tr}(F_sF_s^{T}AR_s^{T}R_sA^{T})$. In the
same way we can show that the fourth and fifth terms in Eq.~(\ref{eq-auxiliary}) equal the fourth and fifth terms in
Eq.~(\ref{eq-aux-h}) respectively when setting $A'=A$.
Now we can show that $Z(A,A)=J(A)$. Thus the conditions of Definition 1 are satisfied.
\end{proof}
Now we try to find the global minimum of $Z(A,A)$. Fixing $A'$, we have
\begin{eqnarray}
\frac{\partial Z(A,A')}{\partial A}=-2F_s\mathcal{P}_sR_s+2\frac{(F_s^{T}F_sA'R_s^{T}R_s)_{(ij)}A_{(ij)}}{A'_{(ij)}} \nonumber \\
-2F_t\mathcal{P}_tR_t+2\frac{(F_t^{T}F_tA'R_t^{T}R_t)_{(ij)}A_{(ij)}}{A'_{(ij)}}+2\frac{(\beta A')_{(ij)}A_{(ij)}}{A'_{(ij)}}.
\label{eq-parital-fs}
\end{eqnarray}
We set $\frac{\partial Z(A,A^{'})}{\partial A}=0$ then we have update rule as follows
\begin{equation}
A_{(ij)} = A_{(ij)}^{'}\frac{(F_s^{T}\mathcal{P}_sR_s+F_t^{T}\mathcal{P}_tR_t)_{(ij)}}{(F_s^{T}F_sA'R_s^{T}R_s+F_t^{T}F_tA'R_t^{T}R_t+\beta A')_{(ij)}}.
\label{eq-parital-fs}
\end{equation}
Further we have
\begin{eqnarray}
\frac{\partial Z(A,A^{'})}{\partial A \partial A} = 2\frac{(F_s^{T}F_sA'R_s^{T}R_s)_{(ij)}}{A'_{(ij)}} \nonumber \\
+2\frac{(F_t^{T}F_tA'R_t^{T}R_t)_{(ij)}}{A'_{(ij)}}+2\frac{(\beta A')_{(ij)}}{A'_{(ij)}}.
\end{eqnarray}
We can show that the second partial derivative is positive. Thus, $Z(A,A^{'})$ is a convex function and we can achieve its global minimum by using Eq.~(\ref{eq-parital-fs}). In other words, we have
$A^{(t+1)}=\arg\min_{A}Z(A,A^{(t)})$ by using our update rule.
The update rule satisfies  Eq.~(\ref{eq-kkt-condition}).

By using the update rule we have
\begin{equation}
J(A^{(t)}) = Z(A^{(t)},A^{(t)}) \geq Z(A^{(t+1)},A^{(t)}) \geq J(A^{(t+1)}),
\end{equation}
where it shows $J(A)$ is monotonically decreasing. Thus the value of $J$ will monotonically decrease under the update rule. The update rule can minimize $J$.

So far we assume others are fixed except $A$. Similarly we can update other variables in the
same way while fixing remaining variables and the update rules are as follows:
\begin{equation}
F_{s(ij)} \gets F_{s(ij)}\frac{(\mathcal{P}_sR_sA^{T})_{(ij)}}{(F_sF_s^{T}\mathcal{P}_sR_sA^{T})_{(ij)}},
\end{equation}
\begin{equation}
R_{s(ij)} \gets R_{s(ij)}\frac{(\mathcal{P}^{T}_sF_sA)_{(ij)}}{(R_sR_s^{T}\mathcal{P}_s^{T}F_sA)_{(ij)}},
\end{equation}
\begin{equation}
F_{t(ij)} \gets F_{t(ij)}\frac{(\mathcal{P}_tR_tA^{T})_{(ij)}}{(F_tF_t^{T}\mathcal{P}_tR_tA^{T})_{(ij)}},
\end{equation}
\begin{equation}
R_{t(ij)} \gets R_{t(ij)}\frac{(\mathcal{P}^{T}_tF_tA)_{(ij)}}{(R_tR_t^{T}\mathcal{P}_t^{T}F_tA)_{(ij)}}.
\end{equation}

We can alternatively update $F_s$, $R_s$, $F_t$, $R_t$
and residue $J(F_s,R_s, F_t, R_t, A)$ will monotonically decrease
\begin{eqnarray}
J(F_s^{(0)},R_s^{(0)}, F_t^{(0)}, R_t^{(0)}, A^{(0)}) \nonumber \\
\geq J(F_s^{(1)},R_s^{(0)}, F_t^{(0)}, R_t^{(0)}, A^{(0)}) \nonumber \\
\geq J(F_s^{(1)},R_s^{(1)}, F_t^{(0)}, R_t^{(0)}, A^{(0)}) \geq ...\nonumber \\
\geq J(F_s^{(1)},R_s^{(1)}, F_t^{(1)}, R_t^{(1)}, A^{(1)}) \geq ...
\end{eqnarray}
Since the lower bound of Eq.~(\ref{eq-opt}) is 0. Our update rules can guarantee convergence.

\subsubsection{Computing $k$}
In most existing works that involve nonnegative matrix factorization, there is a lack of discussions on how to determine the number of features $k$. In our work, we devise a heuristic strategy to optimize the value of $k$, when the objective function Eq.~(\ref{eq-opt}) is optimized to find the common latent structure features.

The goal of learning new structure features is to benefit the classification performance on the target data. To estimate the number of features, an appropriate criterion is that we can measure its ability to represent different classes of the target data. In other words, we want the nodes in the same class to have similar features, yet the nodes belonging to different classes to be separated from each other. Given a specific number $k$ of latent features, we can compute a latent feature space $A$, and accordingly, we have a new feature representation $F_t$ for the nodes in the target network. Given the new feature representation $F_t$ in the target network, we compute a correlation matrix as follows
\begin{equation}
C_k = F_tF_t^{T},
\end{equation}
where element $c_{kij}$ of the matrix $C_k$ represents the similarity between two vectors $v_i$
and $v_j$. The smaller the $c_{kij}$ is, the more similar two vectors $v_i$ and $v_j$ are in the new latent feature space. Therefore, based on the matrix $C_k$, we can compute a quality score $Qs$ using the new feature representation of the labeled data as
\begin{equation}
\mathcal{Q}=\sum_{c=1}^{C}\frac{1}{N_c}\sum_{i,j \in Z_c}c_{kij},
\end{equation}
where $Z_c$ is the set of nodes which belong to class $c$, and $N_c$ is the number of nodes in $Z_c$. This quality score would have a higher value if the nodes in each category are more similar. Therefore, the number of latent feature can be automatically determined by evaluating the local maximum value of this quality score. In summary, our proposed strategy works as follows: given a maximum number of latent features $K$, for $k=2,...,K_{max}$, we compute $A$ by using our algorithm iteratively. We can find the optimal number of latent features such that the corresponding quality score $\mathcal{Q}$ is maximized.

\section{Transfer Learning for ICA}
After discovering the common latent structure features, our next step is to perform collective classification on the target network. Given the target network $G_t=(\mathcal{V}^{u}_t, \mathcal{V}^{l}_t, \mathcal{E}_t)$, we need to train a classifier to predict the labels of the unlabeled nodes $\mathcal{V}^{u}_t$. However, since there only exist a small number of labeled nodes $\mathcal{V}^{l}_t$ in the target domain, we resort to transferring structure features from the source network to facilitate the collective classification task in the target network.

For our classification problem, we adopt an iterative classification algorithm (ICA) that leverages label correlations to predict node labels in the target network. After identifying the common latent feature space $A$, we have new structure features $F_t$ for the target network. These structure features capture the common structure patterns shared by two networks, and thus serve as domain-independent features that are transferred between networks. To capture label correlations in the neighborhood, we also compute relational features by using an aggregation function, such as $\mathtt{count}$, $\mathtt{mode}$, and $\mathtt{proportion}$, to aggregate the label information from the neighbors $\mathcal{N}_i$ of each node $v^{i}_{t}$. By combining node features, structure features, and relational features, we train an ICA classifier that iteratively updates the predictions of all the nodes by using the previous predictions for unknown labels in the neighborhood, until the algorithm converges.

The detailed description of our transfer learning algorithm for collective classification is summarized in Algorithm~\ref{alg-TrCC}.

\begin{algorithm}
\caption{\textbf{Tr}ansfer Learning for \textbf{ICA}}
\label{alg-TrCC}
\begin{algorithmic}[1]
\REQUIRE The source network $G_s=(\mathcal{V}^{l}_{s},\mathcal{E}_s)$ and the target network $G_t=(\mathcal{V}^{u}_t, \mathcal{V}^{l}_t, \mathcal{E}_t)$, a base learning algorithm $f$
\ENSURE Labels of unlabeled nodes in $\mathcal{V}^{u}_t$

\STATE Calculate the label propagation matrix $\mathcal{P}_s$ for $G_s$ and $\mathcal{P}_t$ for $G_t$ using Algorithm \ref{alg-propagation}.

\STATE Calculate the common structure feature space by solving the optimization problem Eq.~(\ref{eq-opt}).

\STATE Reconstruct features of the target data by adding new features $F_t$.


\FOR{each node $v^{i}_{t}$ in $G_t$}

\STATE Compute relational features using only observed nodes in $\mathcal{N}_i$

\STATE Predict the label for an unlabeled node: $y^{i}_{t} \gets f(v^{i}_t)$

\ENDFOR

\WHILE{ All $y^{i}_{t}$'s are not stabilized or number of iterations does not equal a threshold}

\STATE Generate an ordering $\mathcal{O}$ over nodes in $G_t$

\FOR{each node $v_t^{i} \in \mathcal{O}$}

\STATE Compute relational features using the current labels of $\mathcal{N}_i$

\STATE Predict the label for an unlabeled node: $y^{i}_{t} \gets f(v^{i}_t)$

\ENDFOR

\ENDWHILE
\STATE Assign the last predicted labels to $\mathcal{V}^{u}_t$
\end{algorithmic}
\end{algorithm}

\section{Experiments}
To evaluate the performance of our proposed algorithm, we perform extensive experiments on four real-world networks.

\subsection{Data sets}
The four real-world data sets used in our experiments include: CiteSeer, Cora, WebKB and Terrorist Attacks\footnote{\url{http://www.cs.umd.edu/projects/linqs/projects/lbc/index.html}}.
For the data sets, we ignore the node's self-links and the direction of links, and thus two nodes are connected if either of them has a directed link to the other. In the four networks, the features of nodes are different in the domains and the label spaces are also different indicating different classification problems. The detailed description of the four data sets is discussed as follows.

\textbf{CiteSeer:}
The CiteSeer data set consists of 3312 scientific publications and 4732 citation links. Each node is represented by a 0/1-valued word vector indicating absence/presence of the corresponding words from a dictionary of 3703 words, and is labeled
as one of six classes: \emph{Databases (DB)}, \emph{Machine Learning (ML)}, \emph{Information Retrieval (IR)}, \emph{Artificial Intelligence (AI)}, \emph{Human Computer Interaction (HCI)}, and \emph{Agents}. We consider a binary classification problem which takes DB as the positive class and the rest as the negative class.

\textbf{Cora:}
The Cora data set contains 2708 scientific publications classified into one of seven classes: \emph{Probabilistic Methods}, \emph{Neural Networks}, \emph{Case Based}, \emph{Rule Learning}, \emph{Reinforcement Learning}, \emph{Genetic Algorithms} and \emph{Theory}. The citation network contains 5429 links. We consider a binary classification problem and use Neural Networks as the positive class and all others are treated as the negative class.

\textbf{WebKB:}
The WebKB data set contains information about Web pages and their hyperlinks. We use Wisconsin data which contains 265 Web pages and 479 hyperlink relationships. Each Web page is classified into one of five classes: \emph{student}, \emph{course}, \emph{faculty}, \emph{project} and \emph{staff}. We consider the majority class \emph{student} as positive and the rest as negative.

\textbf{Attack:}
This data set consists of 645 terrorist attacks each assigned one of six labels, indicating the type of the attack, including \emph{Bombing}, \emph{Weapon Attack}, \emph{Kidnapping}, \emph{Arson}, \emph{NBCR Attach}, and \emph{Other Attack}. Each node represents a terrorist attack and a link is created between two co-located attacks. Each attack is described by a 0/1-valued vector of attributes whose entries indicate the absence/presence of a feature. There are a total of 106 distinct features. We also take the majority class \emph{Bombing} as positive and the rest as negative.

\begin{table}
\begin{small}
\begin{center}
\begin{tabular}{l c c c c}
\toprule
\textbf{Data Set} & \textbf{CiteSeer} & \textbf{Cora} & \textbf{WebKB} & \textbf{Attack}  \\
\midrule
\# of Nodes & 3312 & 2708 & 265 & 645 \\
\# of Links & 4732  & 5429 & 479 & 3172\\
\# of Classes & 6 & 7 & 5 & 6\\
\# in Largest Class & 701 & 818 & 122 & 312\\
\# in Smallest Class  & 249 & 180 & 22 & 4\\
\bottomrule
\end{tabular}
\end{center}
\end{small}
\caption{Summary of the four data sets}
\label{tb-data-sets}
\end{table}

\subsection{Baselines}
Our proposed algorithm is referred to as \textbf{TrICA} in the experiments. Since our work is the first to perform transfer learning across networks for predicting node labels, and no existing state-of-the-art transfer learning method is available for comparison, we compare \textbf{TrICA} with other two non-transfer-learning baseline methods, with the objective to demonstrate that carefully transferring knowledge from other networks can indeed help improve the node classification accuracy.

\begin{itemize}
\item \textbf{ICA:} This method uses the content features of the labeled nodes in the target network to train an ICA classifier for predicting unlabeled nodes \cite{sen2008collective}.
\item \textbf{Propagation-based ICA (PICA):} This method also relies on the target network to perform collective classification. In addition to the nodes' content features, it also uses a propagation matrix constructed in the target network as structure features to train an ICA classifier.
\end{itemize}

It is worth noting that we have indeed considered to use TrAdaBoost, which is a popular transfer learning algorithm~\cite{Dai2007Boosting}, as a baseline. However, this algorithm assumes that the source and target domains share the same feature space and label space. In contrast, in our problem, the features of the nodes in different networks can be largely different. For example, the feature space of the nodes in CiteSeer contains word occurrences in scientific publications in computer science area, which differs radically from the feature space in Attack where node features represent attributes of attacks. Therefore, TrAdaBoost cannot be used as a baseline to compare with the proposed algorithm.

\subsection{Experimental settings}
In our experiments, we focus on binary classification problems in the target network, in which the largest class for each data set is considered as the positive class, and the rest belongs to the negative class. In the target network, we randomly select a fixed percentage $p$ of nodes as labeled data, and our objective is to build a classifier to predict labels of unlabeled nodes in the network.

For this purpose, we use logistic regression as a base classifier to perform collective classification in the target network. Specifically, we train an ICA classifier that uses $\mathtt{proportion}$ as the aggregation function to compute relational features, which are the proportions of each class in the neighbors of a node to aggregate the label information from the neighbors of each node. Thereafter, the ICA is trained based on a combined set of  aggregated features and other features, depending on the algorithm itself. We apply ICA iteratively to the whole target network until it converges. We then evaluate the classification accuracy only on the unlabeled nodes. For evaluation, we repeat each algorithm for three times and report the average results.


\subsection{Classification performance}
To provide comprehensive validations for transfer learning tasks, we take turns to consider each single data set as the target network and the other three as the source networks, respectively. We perform the first set of experiments to compare the classification accuracy of different methods with respect to different numbers of labeled nodes in the target network. We vary the percentage of labeled nodes $p$ in the target network (from $2\%$ to $60\%$) and run ICA algorithms on the respective data sets. A better classification algorithm is expected to achieve a higher classification accuracy given a same number of labeled data.

\begin{figure*}[!ht]
\centering
\subfigure[\emph{T:CiteSeer-S:Cora}]{\includegraphics[width=0.28\textwidth]{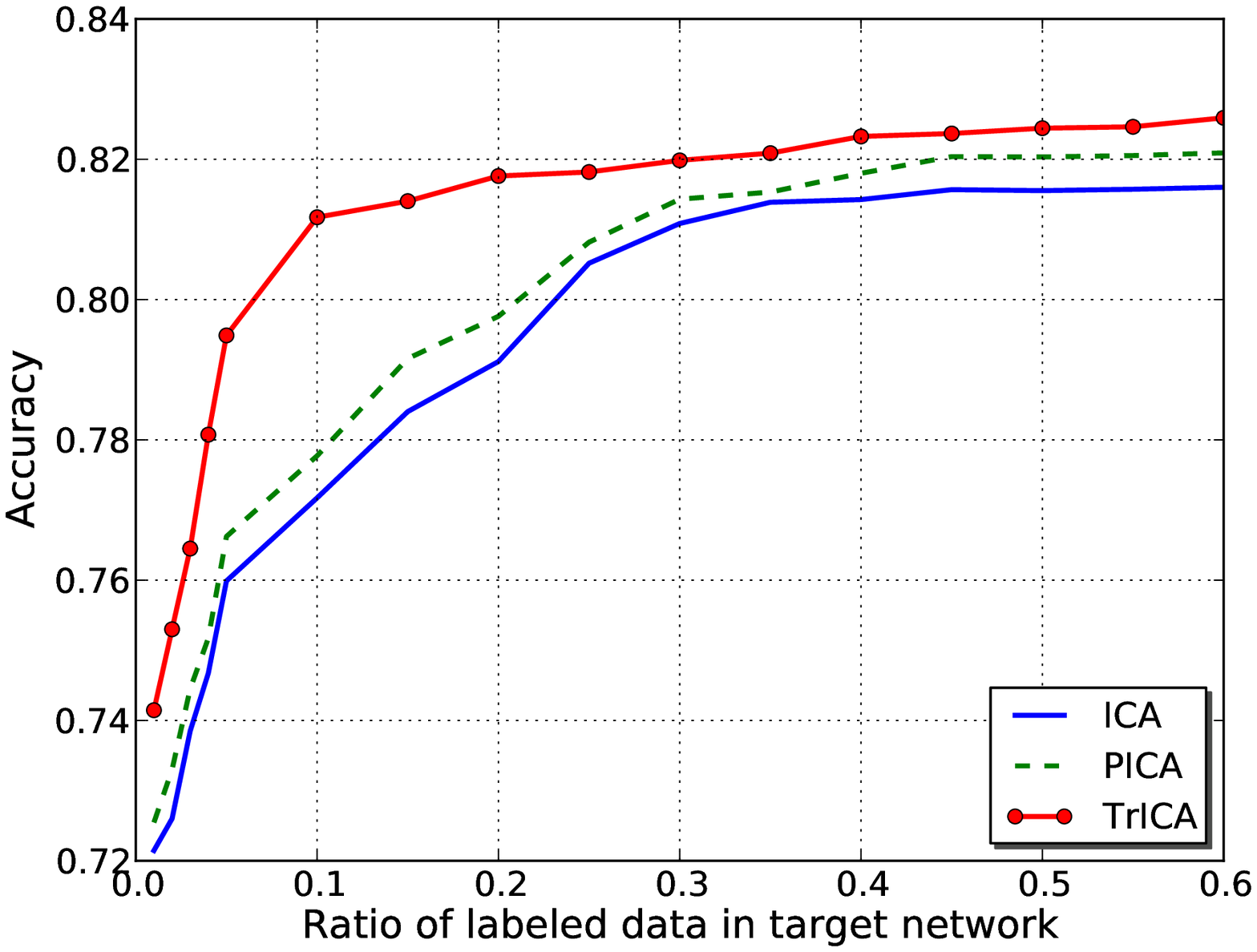}}
\subfigure[\emph{T:CiteSeer-S:WebKB}]{\includegraphics[width=0.28\textwidth]{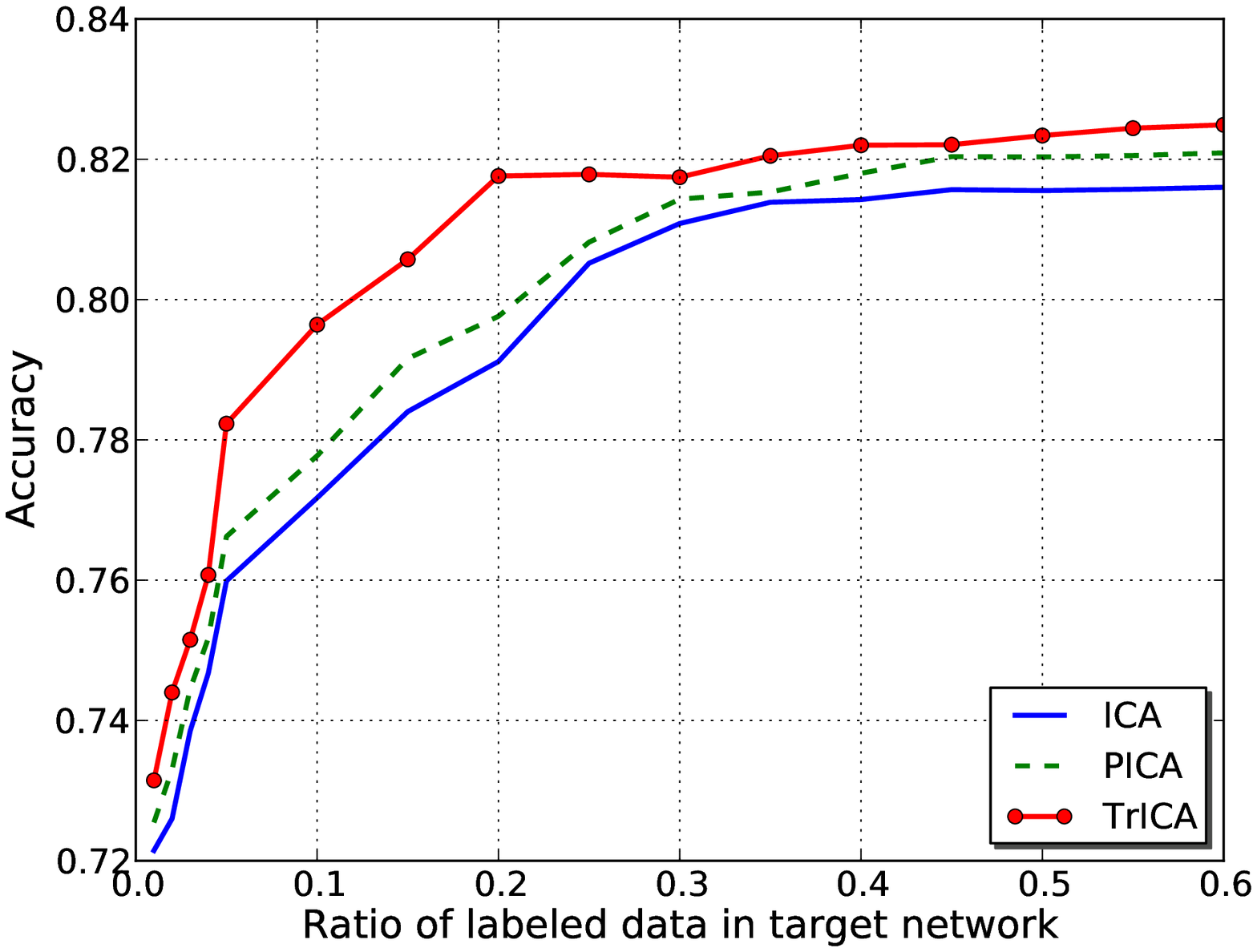}}
\subfigure[\emph{T:CiteSeer-S:Attack}]{\includegraphics[width=0.28\textwidth]{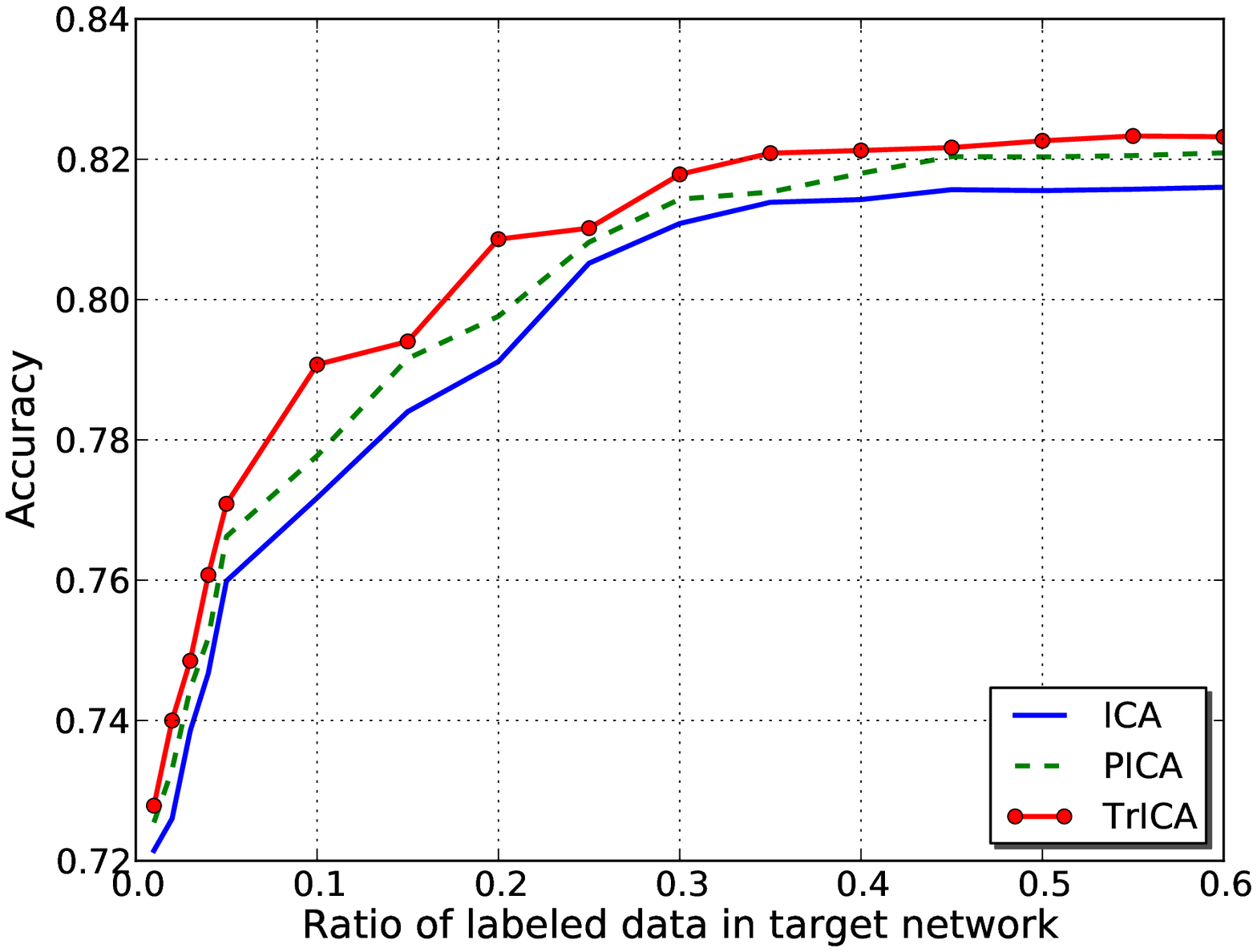}} \\
\subfigure[\emph{T:Cora-S:CiteSeer}]{\includegraphics[width=0.28\textwidth]{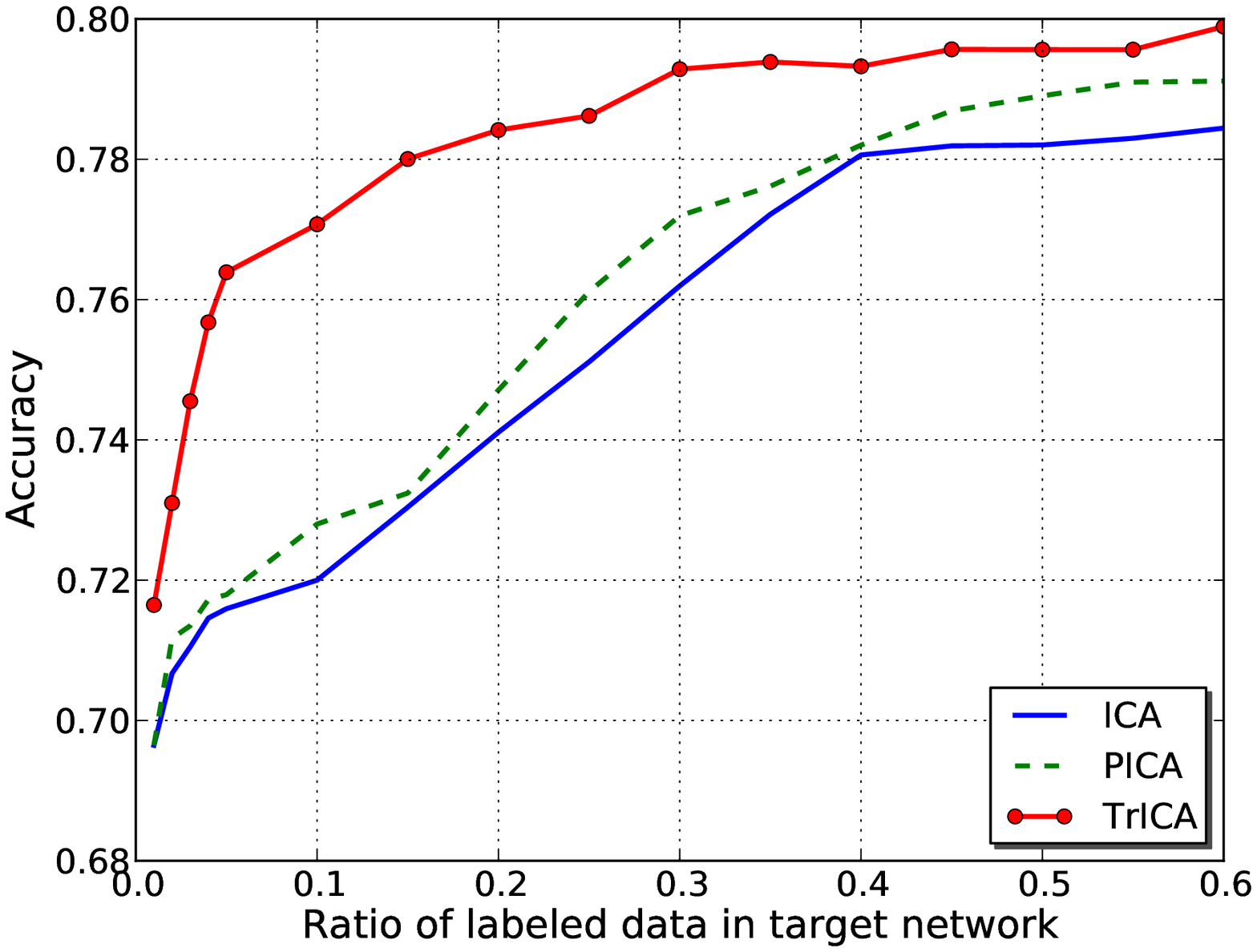}}
\subfigure[\emph{T:Cora-S:WebKB}]{\includegraphics[width=0.28\textwidth]{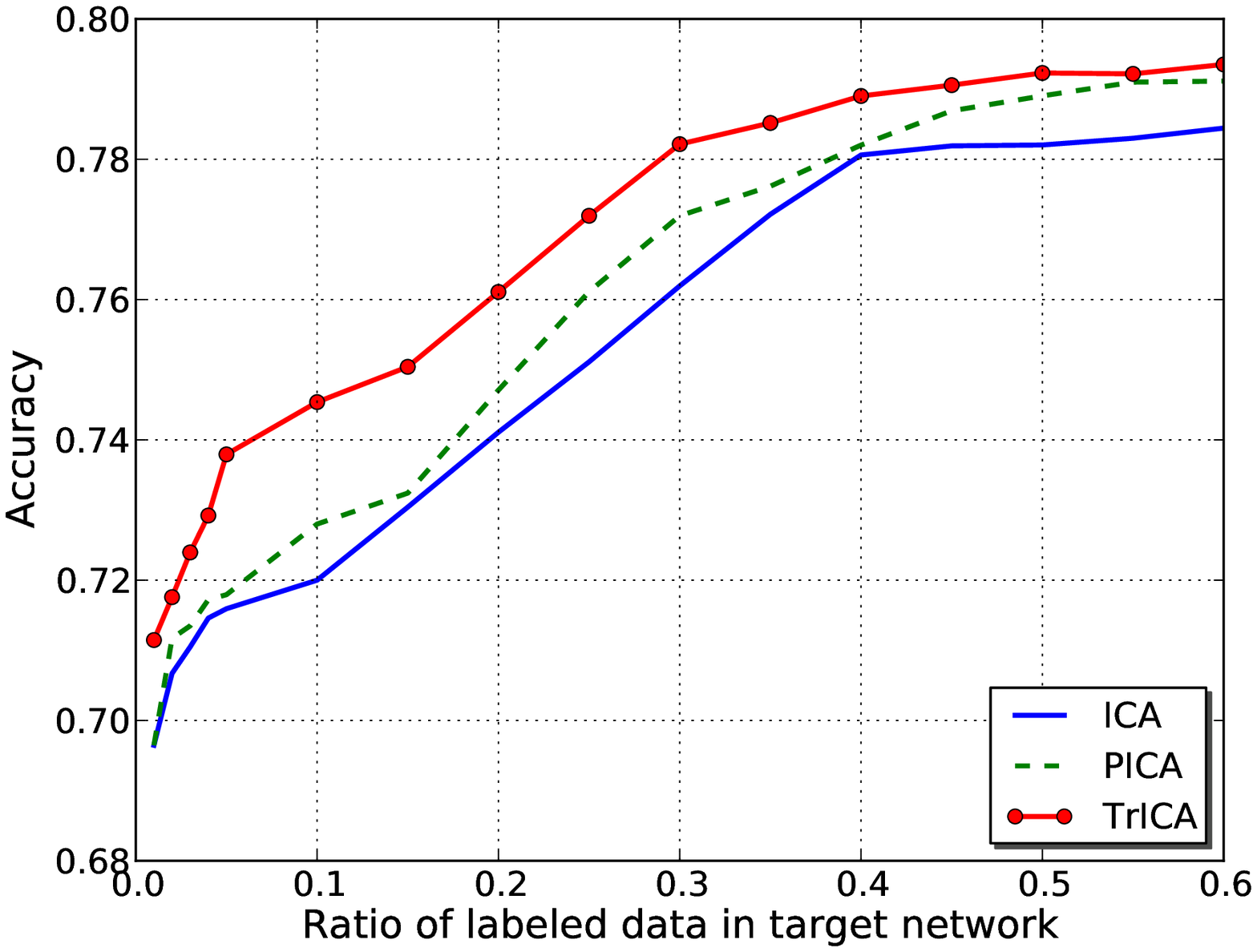}}
\subfigure[\emph{T:Cora-S:Attack}]{\includegraphics[width=0.28\textwidth]{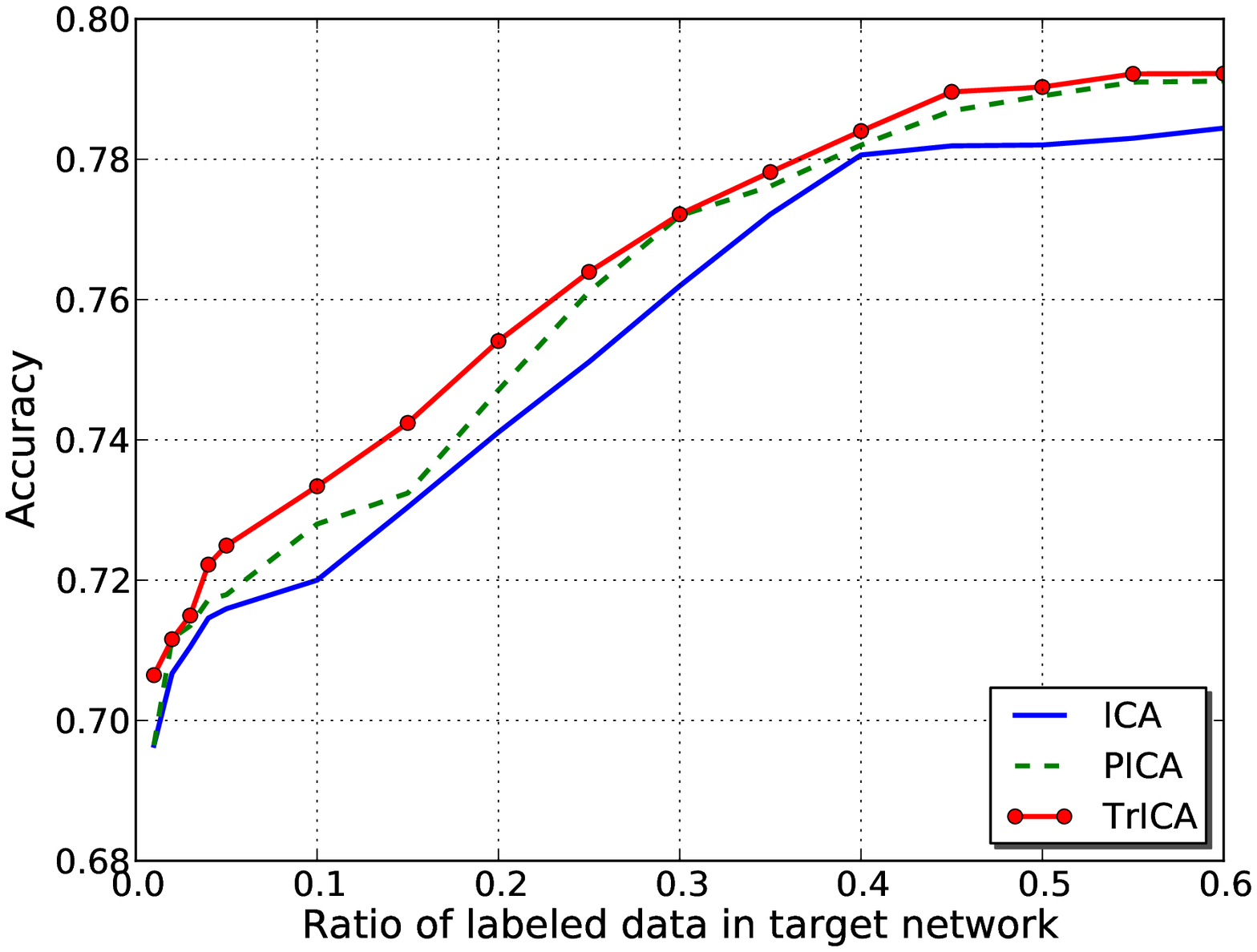}}
\\
\subfigure[\emph{T:WebKB-S:CiteSeer}]{\includegraphics[width=0.28\textwidth]{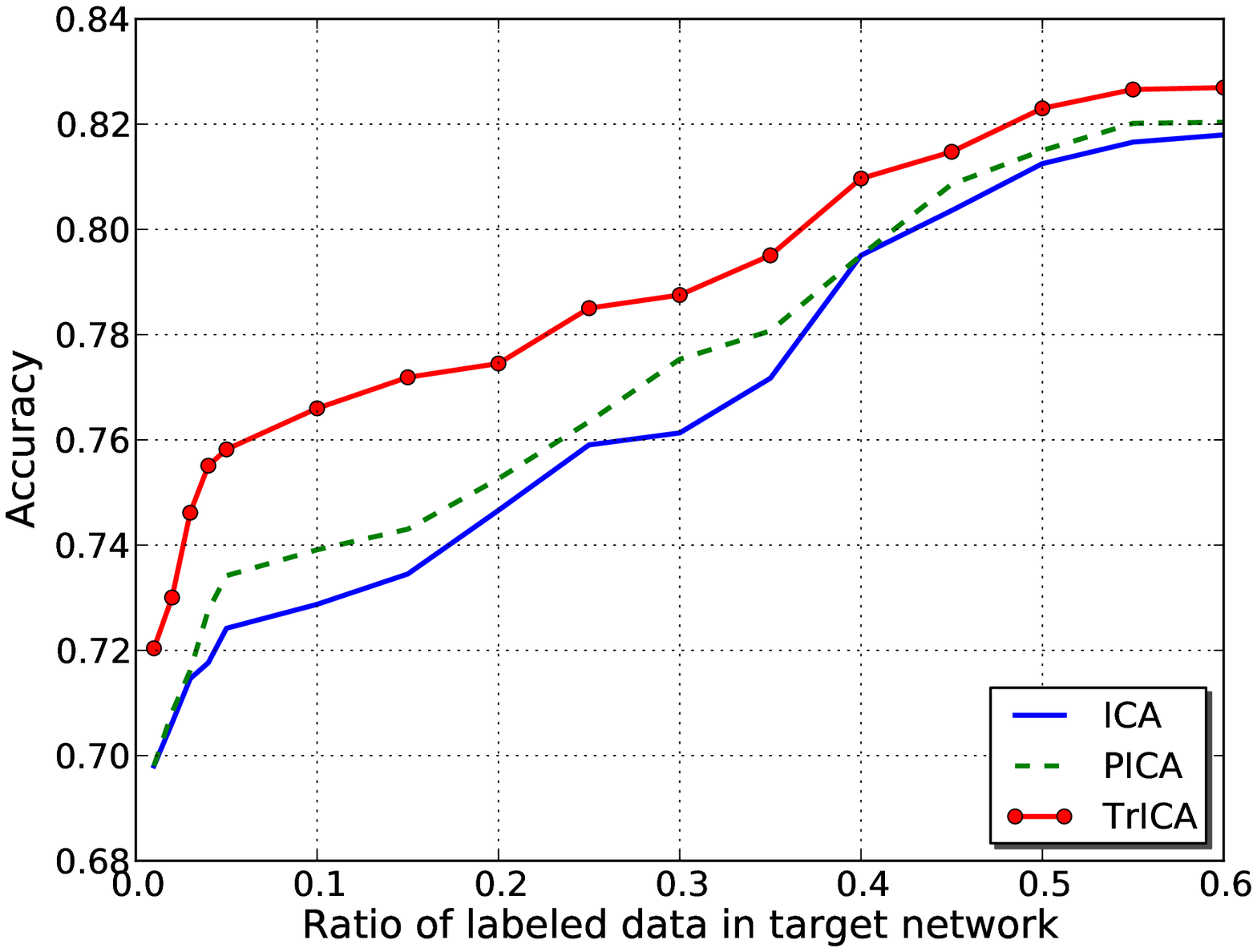}}
\subfigure[\emph{T:WebKB-S:Cora}]{\includegraphics[width=0.28\textwidth]{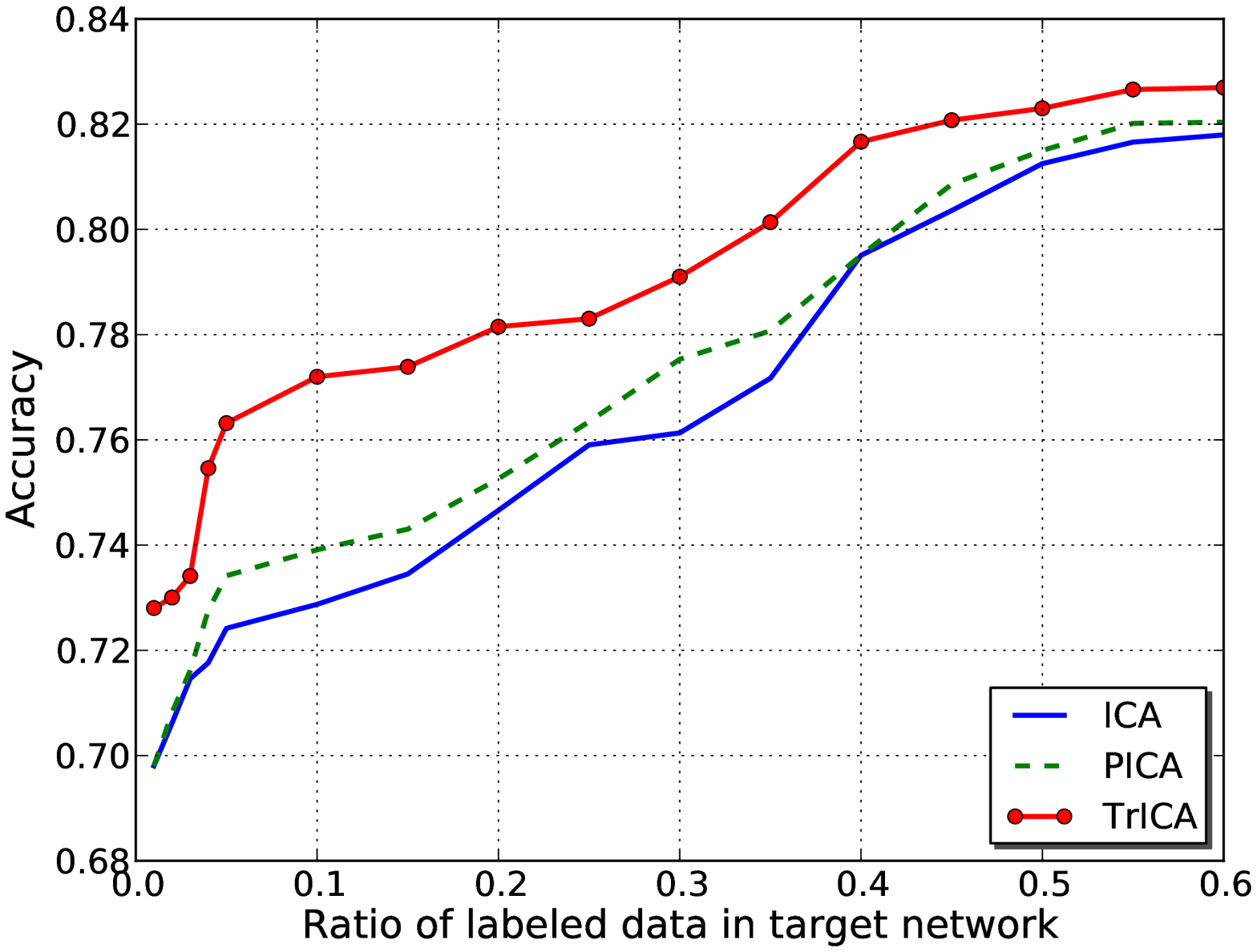}}
\subfigure[\emph{T:WebKB-S:Attack}]{\includegraphics[width=0.28\textwidth]{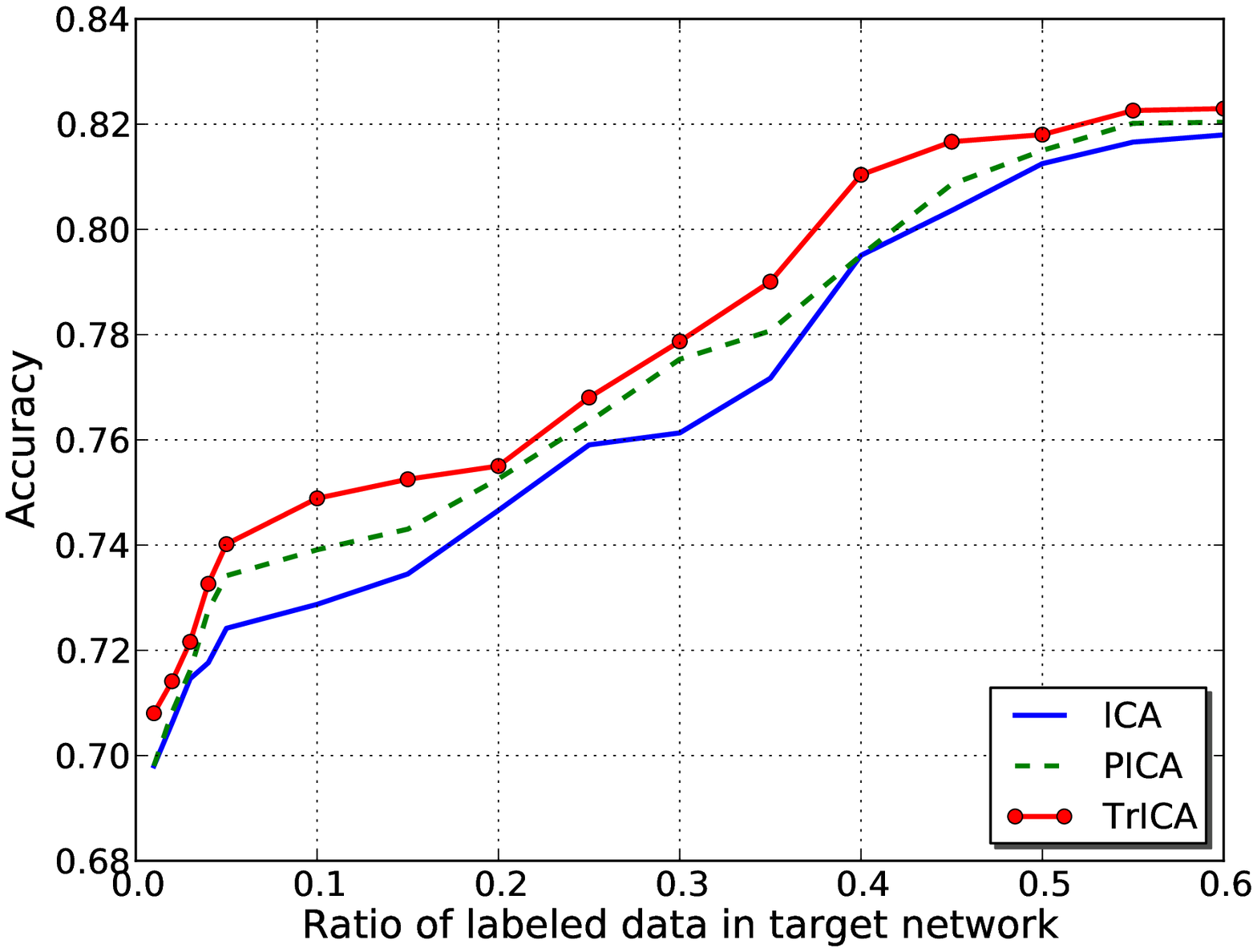}}
\\
\subfigure[\emph{T:Attack-S:CiteSeer}]{\includegraphics[width=0.28\textwidth]{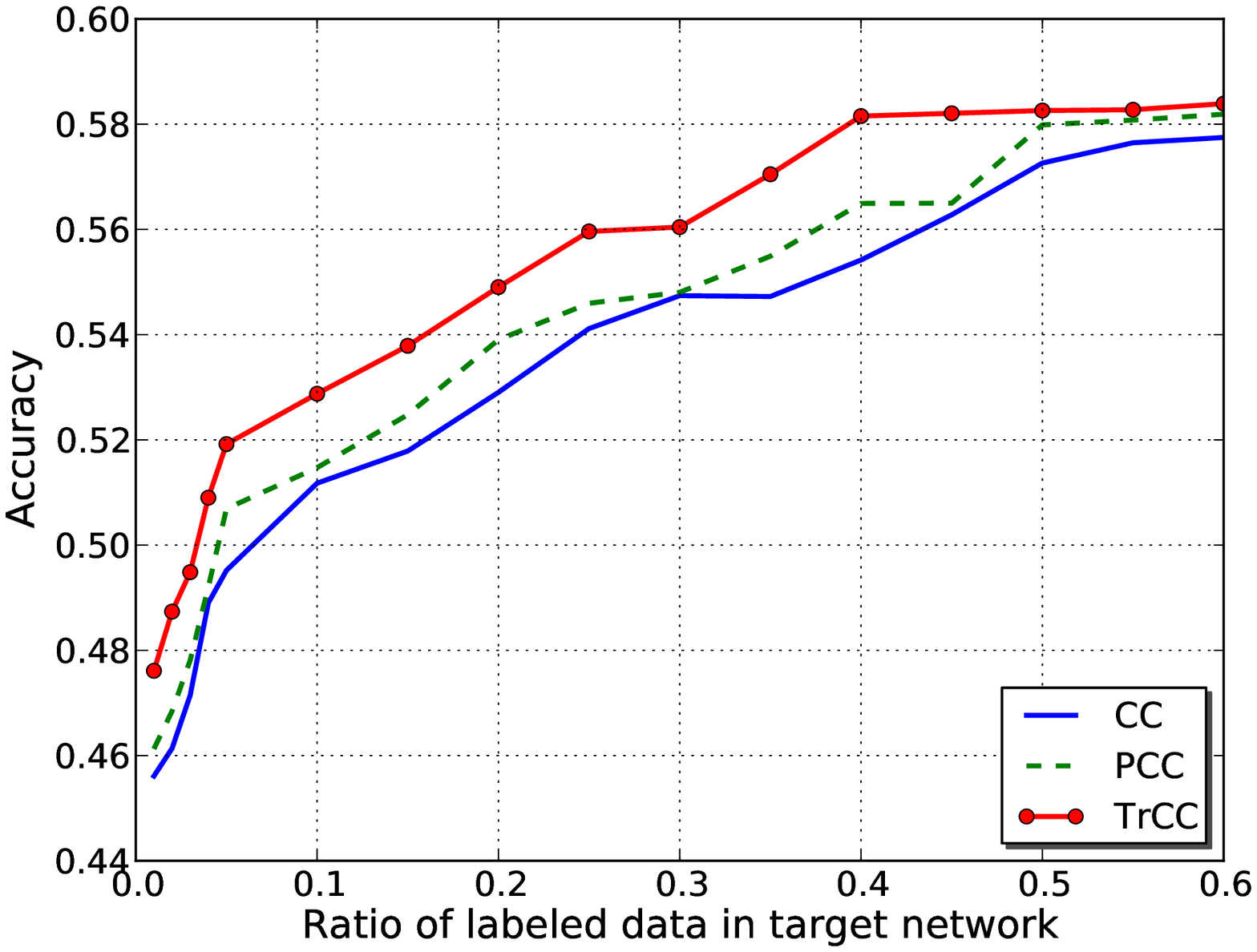}}
\subfigure[\emph{T:Attack-S:Cora}]{\includegraphics[width=0.28\textwidth]{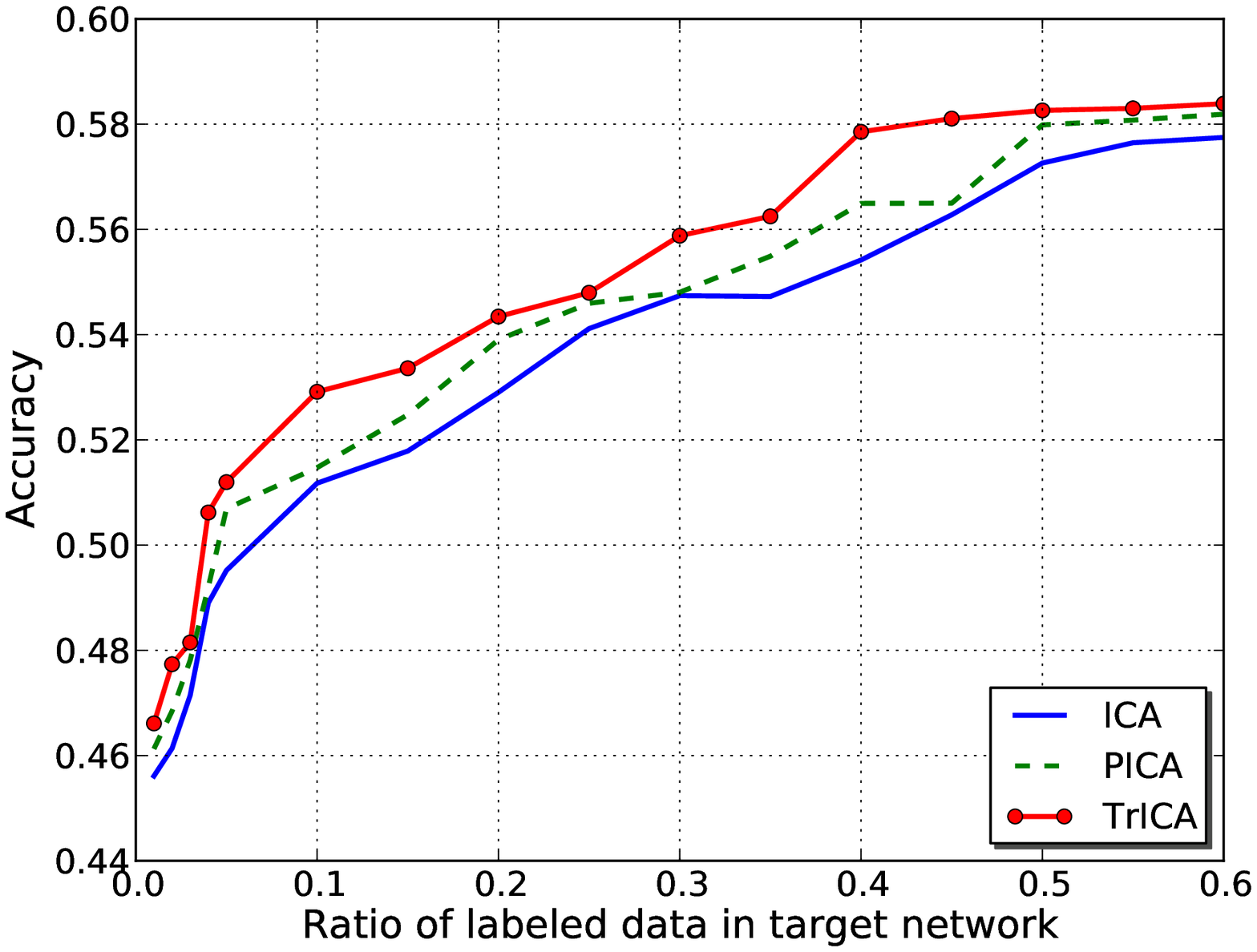}}
\subfigure[\emph{T:Attack-S:WebKB}]{\includegraphics[width=0.28\textwidth]{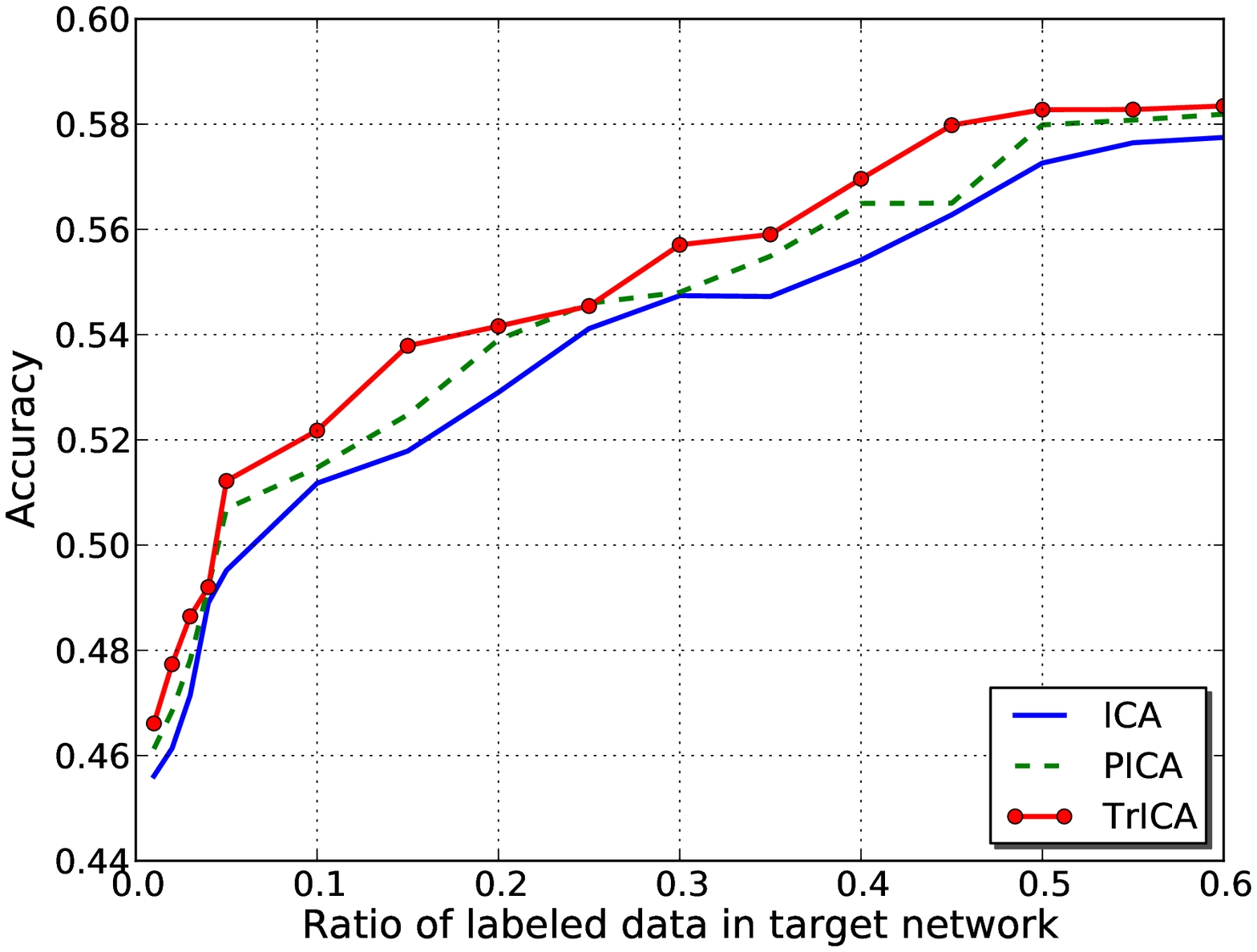}}
\caption{Accuracy comparison of different algorithms on four data sets with respect to different percentages of the labeled nodes in target networks. $T$ indicates target networks and $S$ indicates source networks.}
\label{fig-accuray}
\end{figure*}

The classification results are reported in Figure~\ref{fig-accuray}, which shows that the proposed algorithm TrICA consistently achieves higher accuracy than other baselines over all the transfer learning settings. This confirms that transferring latent structure features across networks can significantly improve the accuracy of classifying nodes in the target network. Noticeably, when CiteSeer is used as the source network and Cora is the target network, or vice versa, TrICA outperforms other baselines to a larger margin, especially when there exists only a small number of labeled nodes in the target network. This is because CiteSeer and Cora are in two similar domains, and they both represent citation relationships between scientific publications. Thus, the two networks share striking similarity in their latent features, which enables transfer learning to be more effective. Meanwhile, PICA is observed to have a better performance than ICA. This indicates that, structure features, discovered via constructing the label propagation matrix, can help improve the collective classification accuracy.

\begin{figure*}[ht]
\centering
\subfigure[\emph{T:CiteSeer}]{\includegraphics[width=0.24\textwidth]{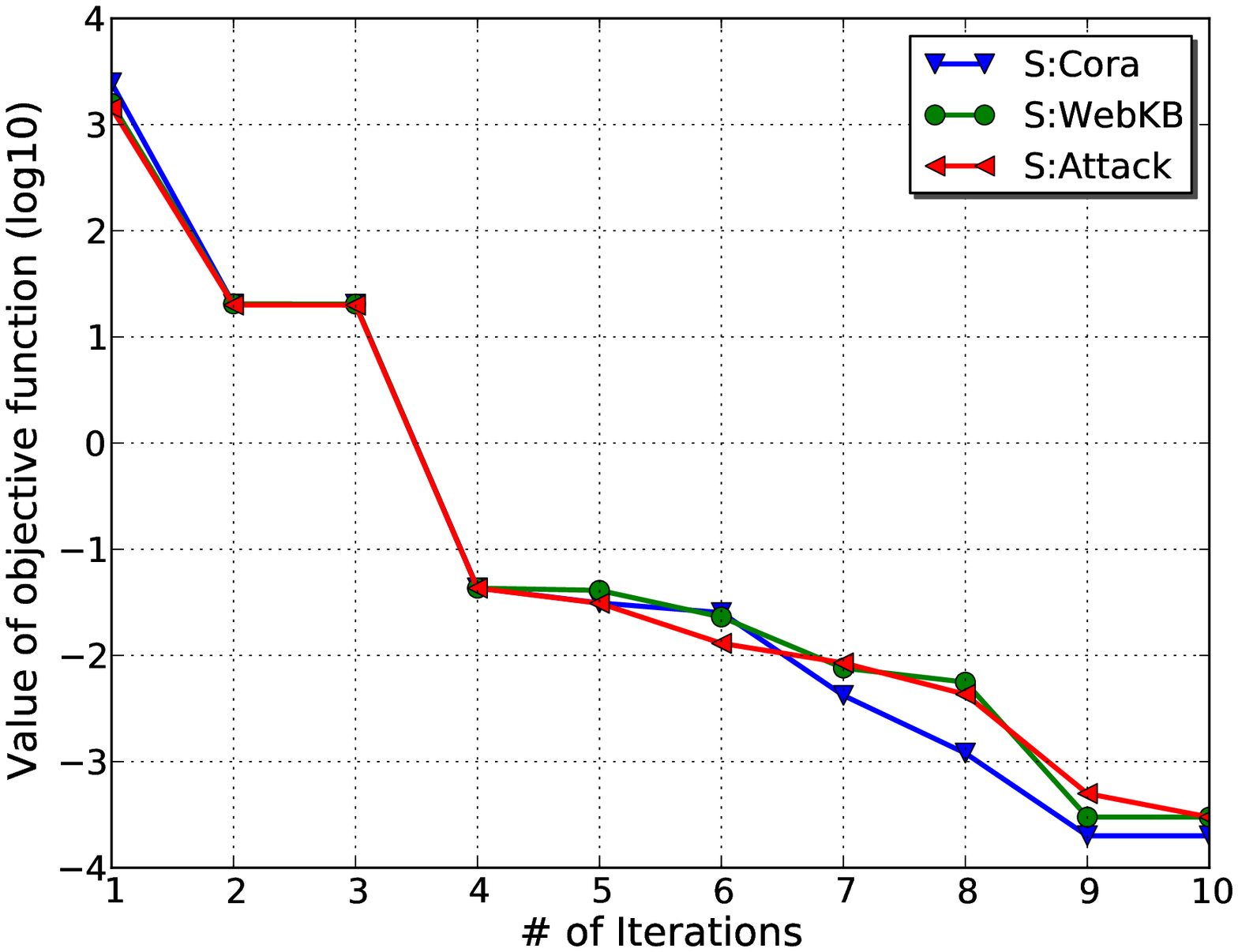}}
\subfigure[\emph{T:Cora}]{\includegraphics[width=0.24\textwidth]{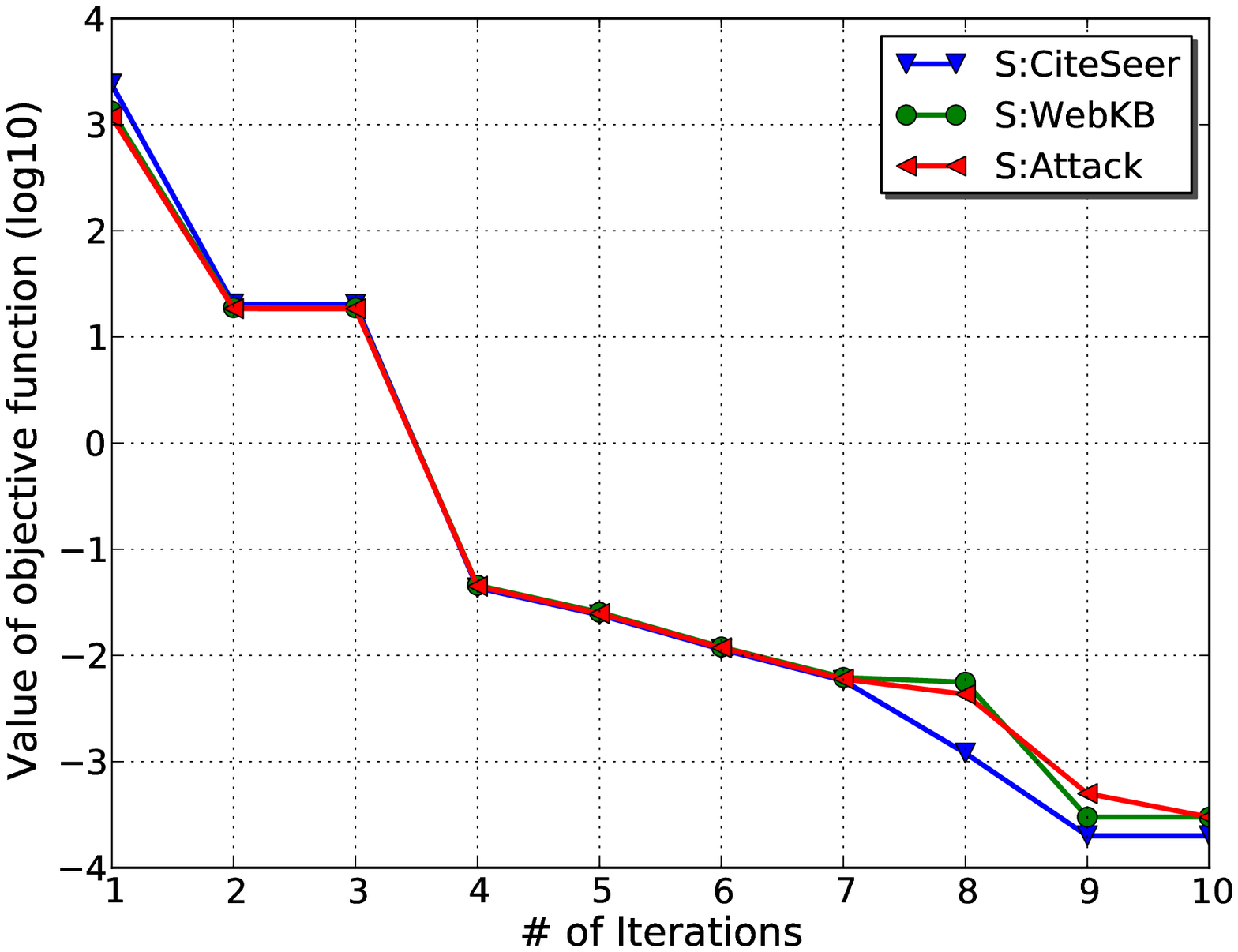}}
\subfigure[\emph{T:WebKB}]{\includegraphics[width=0.24\textwidth]{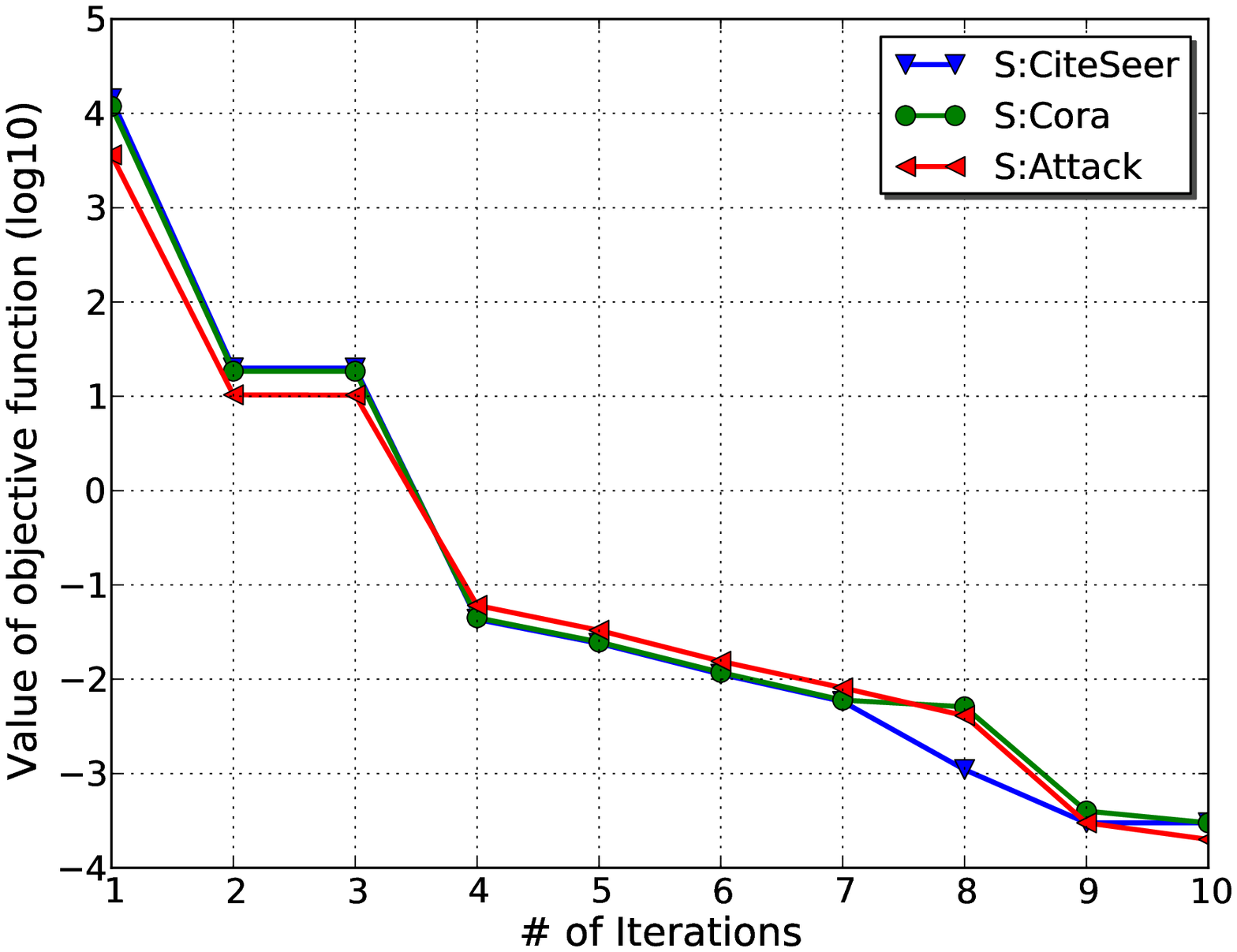}}
\subfigure[\emph{T:Attack}]{\includegraphics[width=0.24\textwidth]{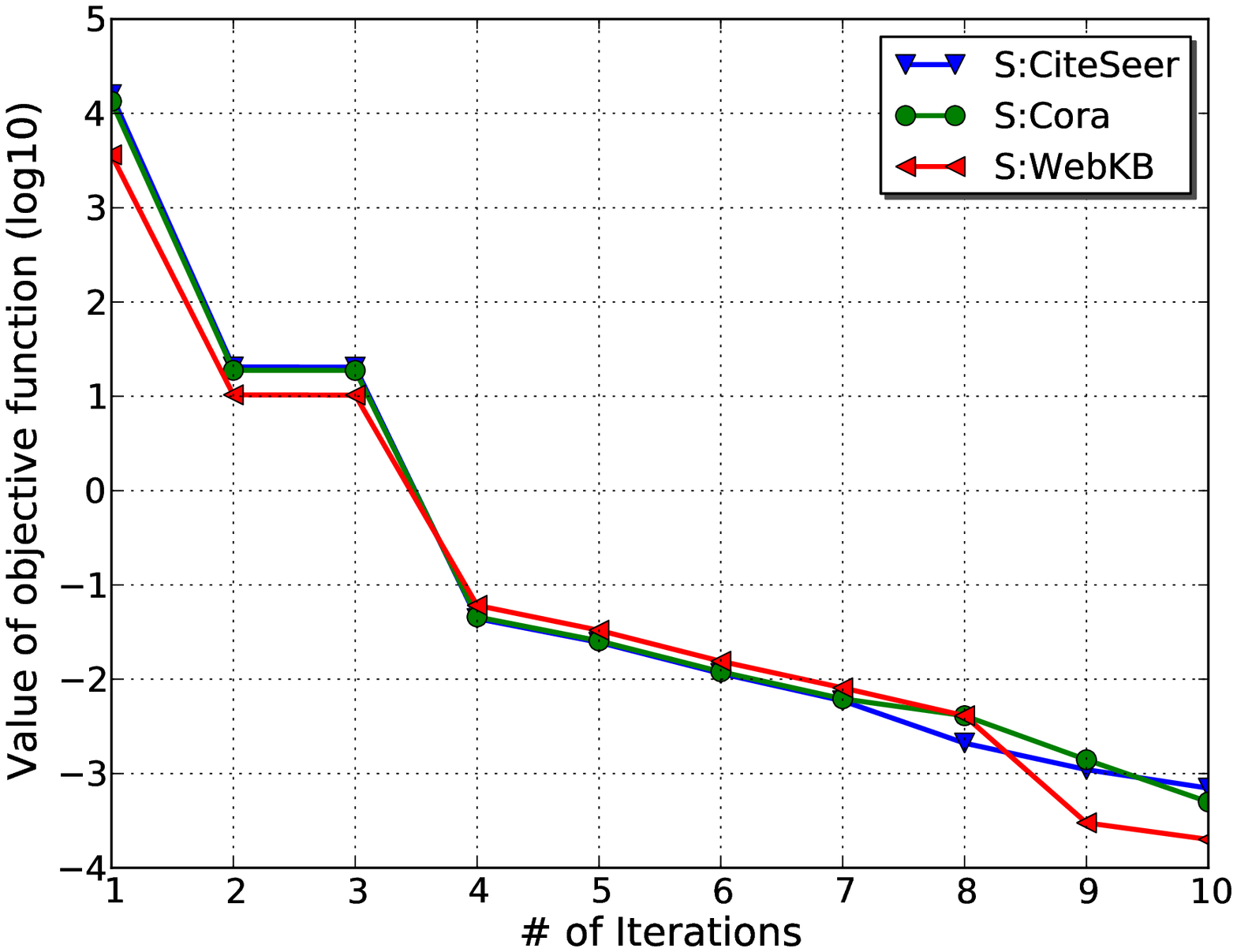}}
\caption{Convergence of the objective function for discovering the common latent structure features. y-axis denotes the value of objective function in $log$ scale and x-axis denotes the number of iterations. T indicates target networks and S indicates source networks. }
\label{fig-convergence}
\end{figure*}

\begin{figure*}[ht]
\centering
\subfigure[\emph{T:CiteSeer}]{\includegraphics[width=0.24\textwidth]{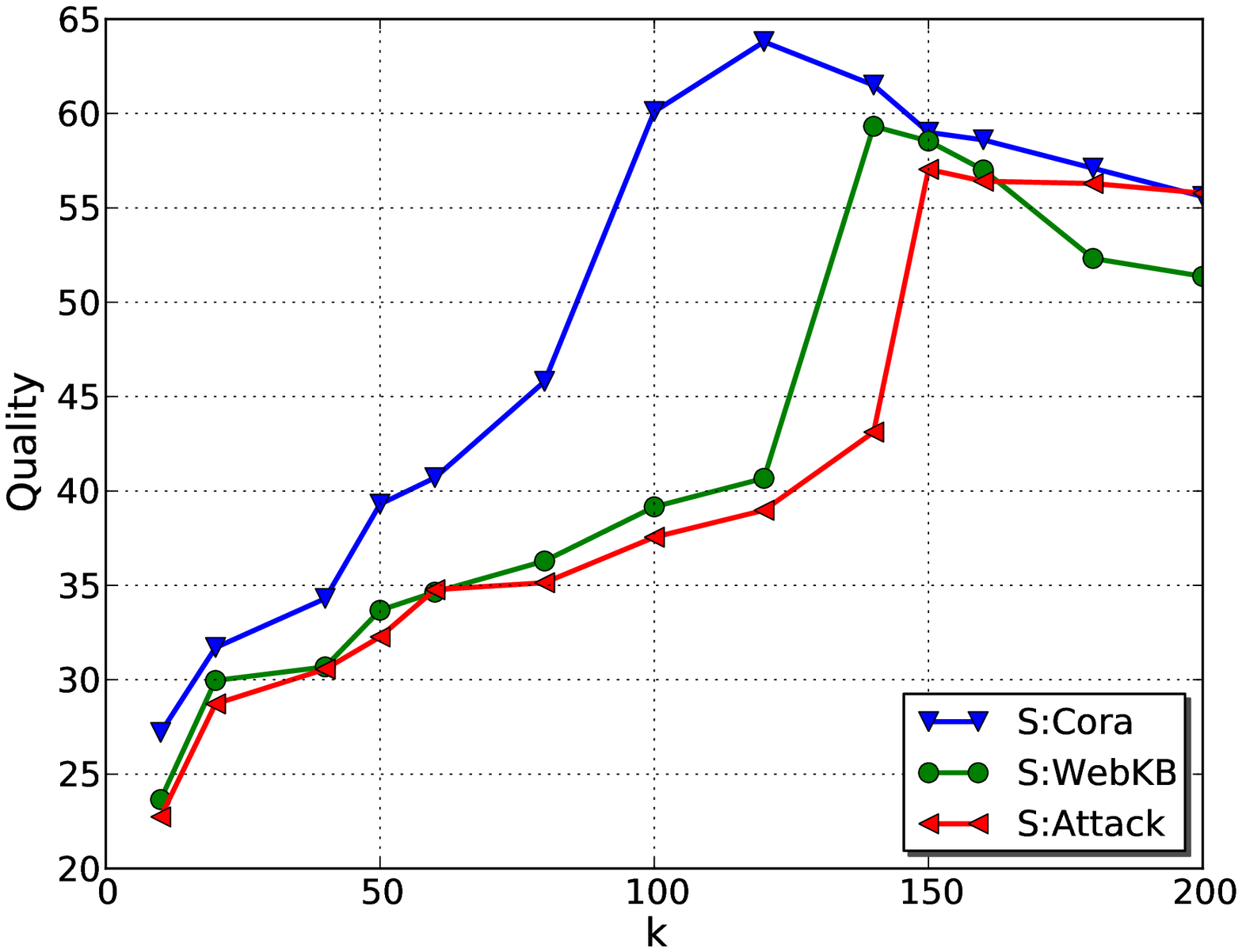}}
\subfigure[\emph{T:Cora}]{\includegraphics[width=0.24\textwidth]{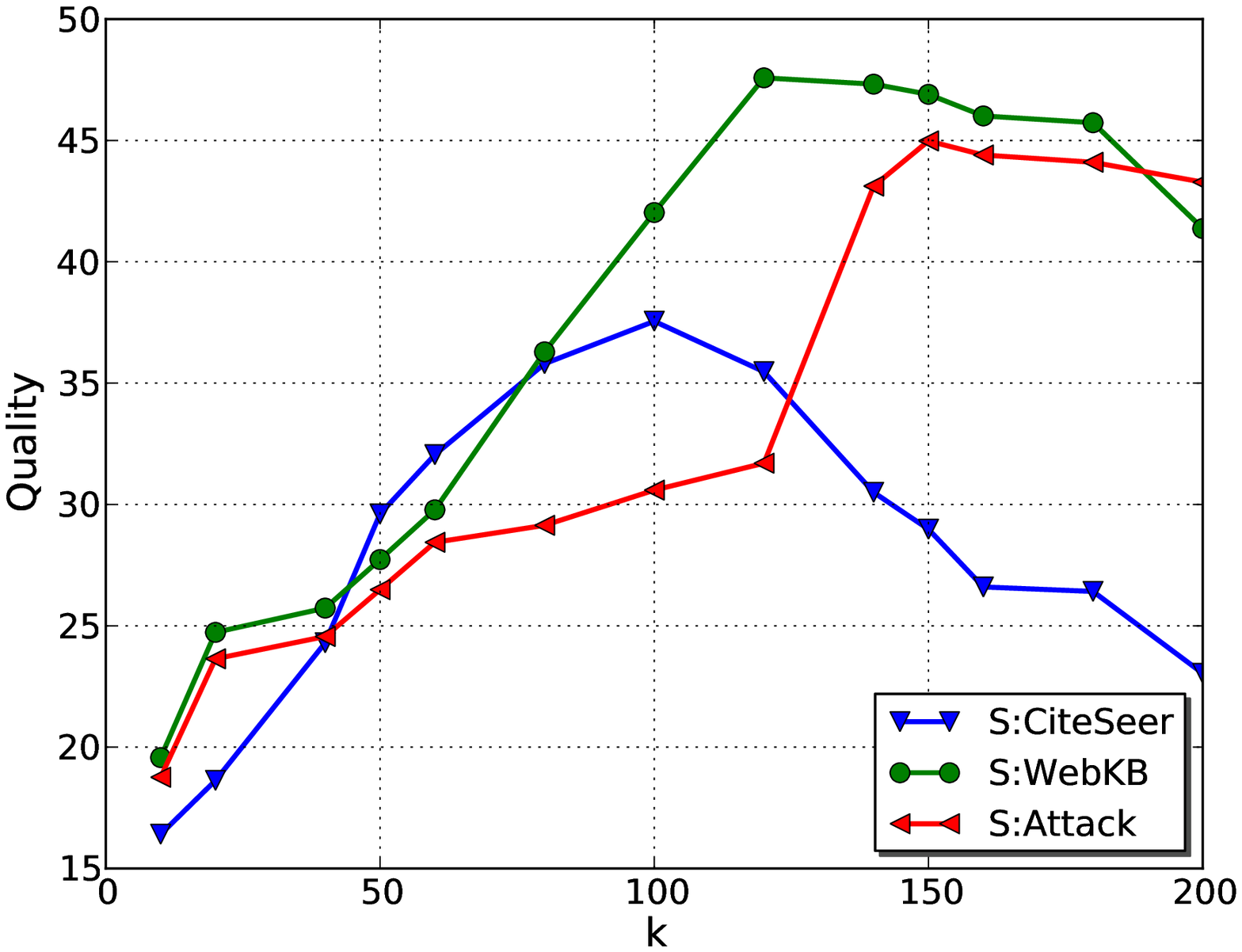}}
\subfigure[\emph{T:WebKB}]{\includegraphics[width=0.24\textwidth]{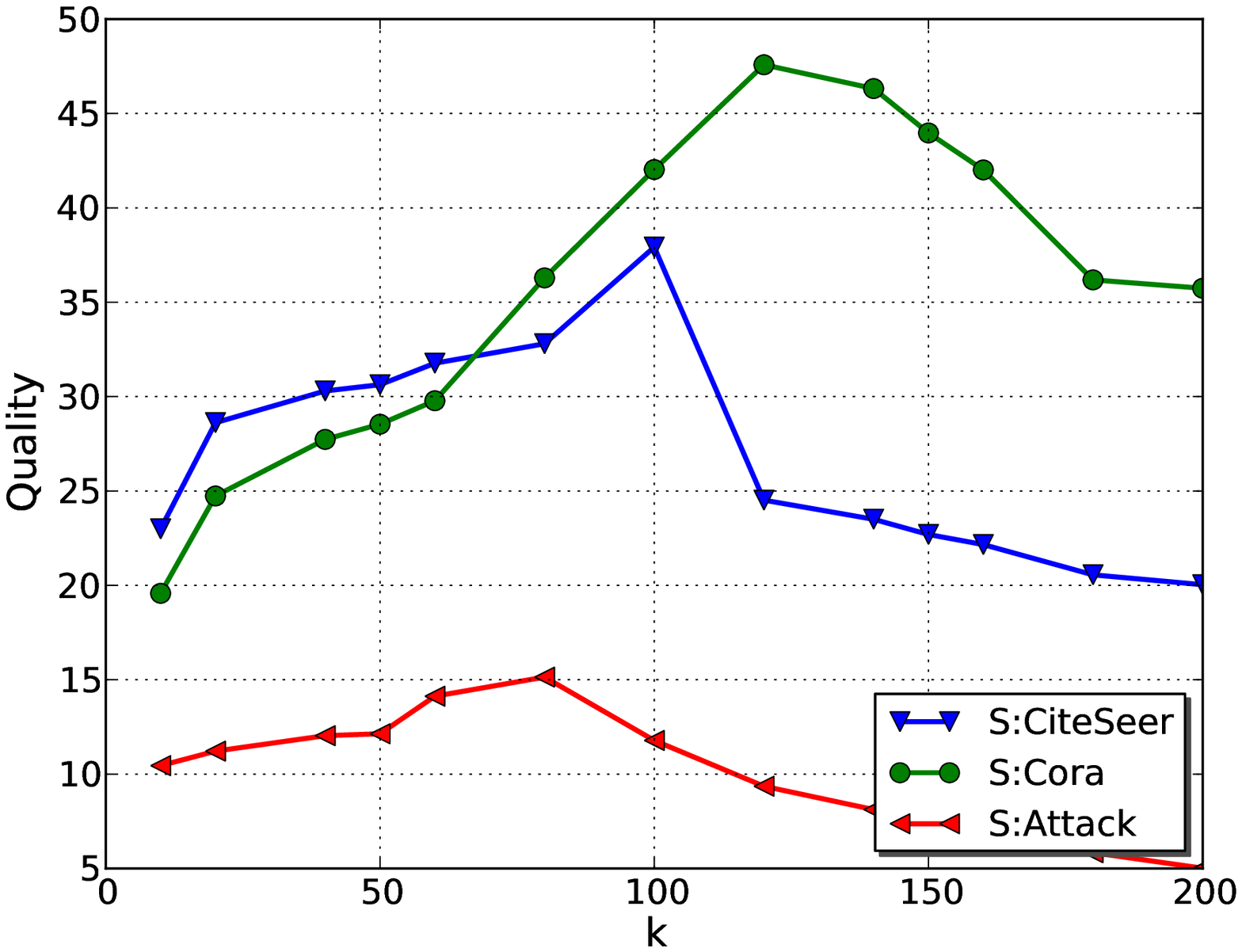}}
\subfigure[\emph{T:Attack}]{\includegraphics[width=0.24\textwidth]{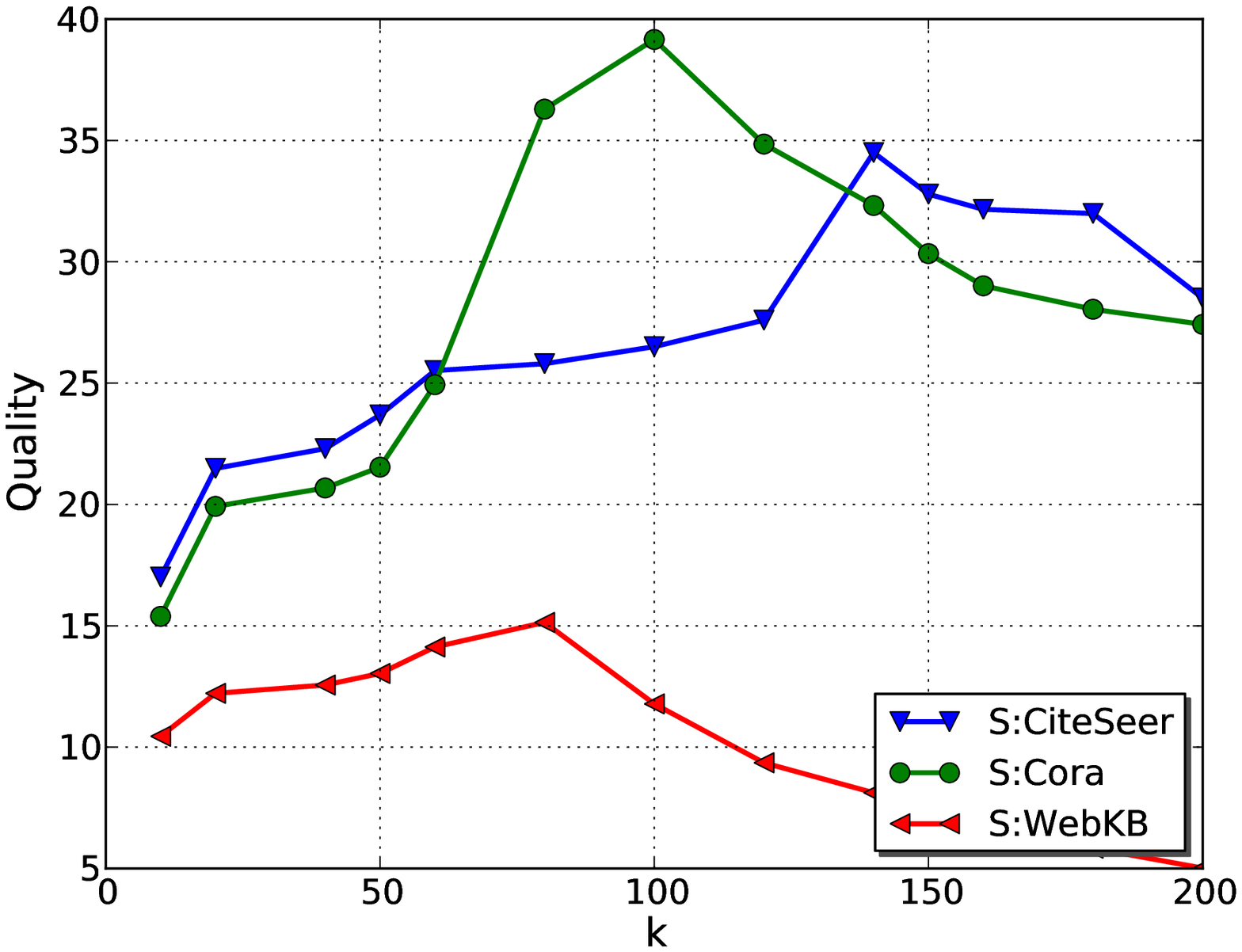}}
\caption{Quality scores with respect to different values of $k$. T indicates target networks and S indicates source networks.}
\label{fig-quality}
\end{figure*}
\begin{figure*}[ht]
\centering
\subfigure[\emph{T:CiteSeer}]{\includegraphics[width=0.24\textwidth]{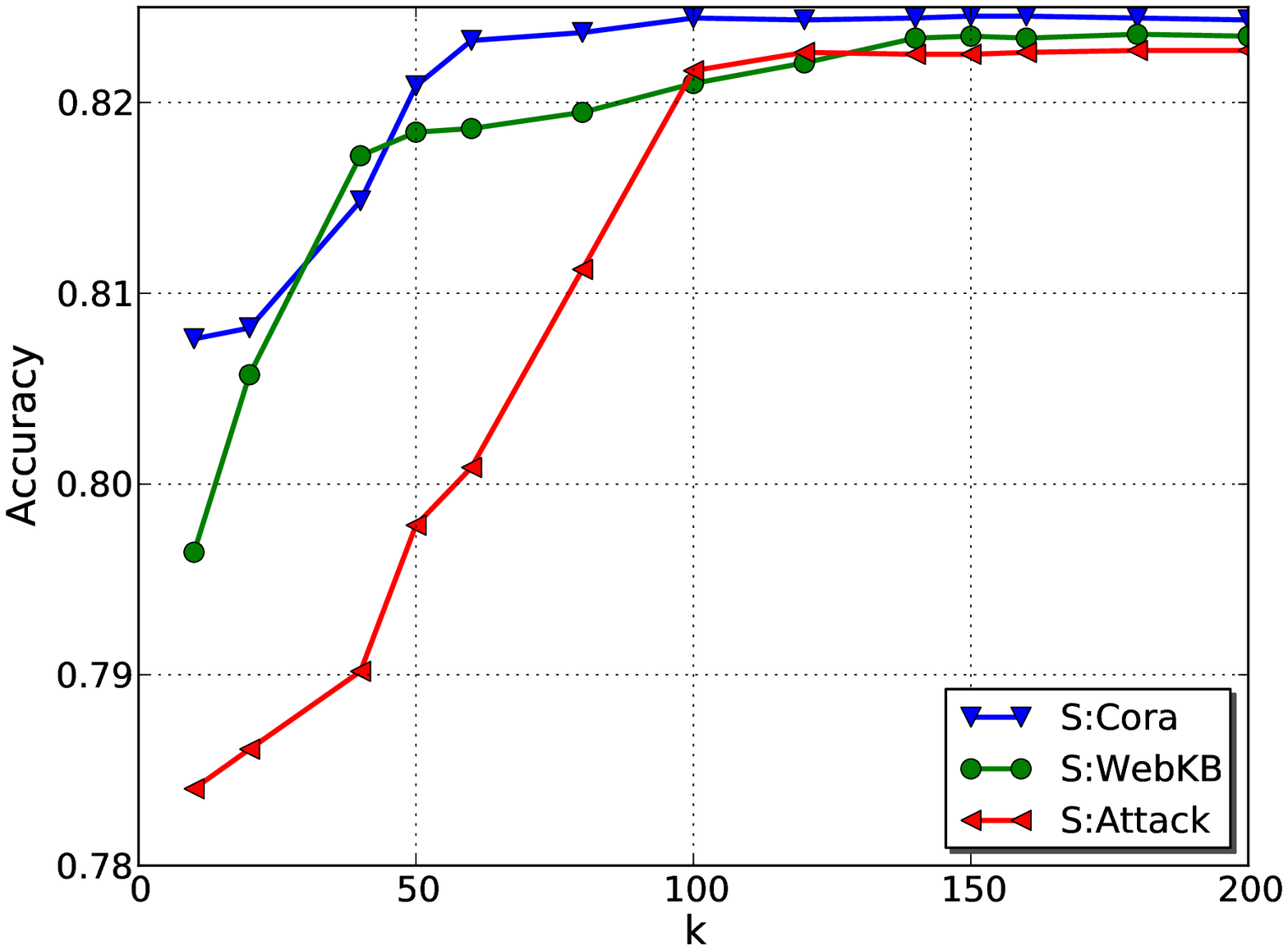}}
\subfigure[\emph{T:Cora}]{\includegraphics[width=0.24\textwidth]{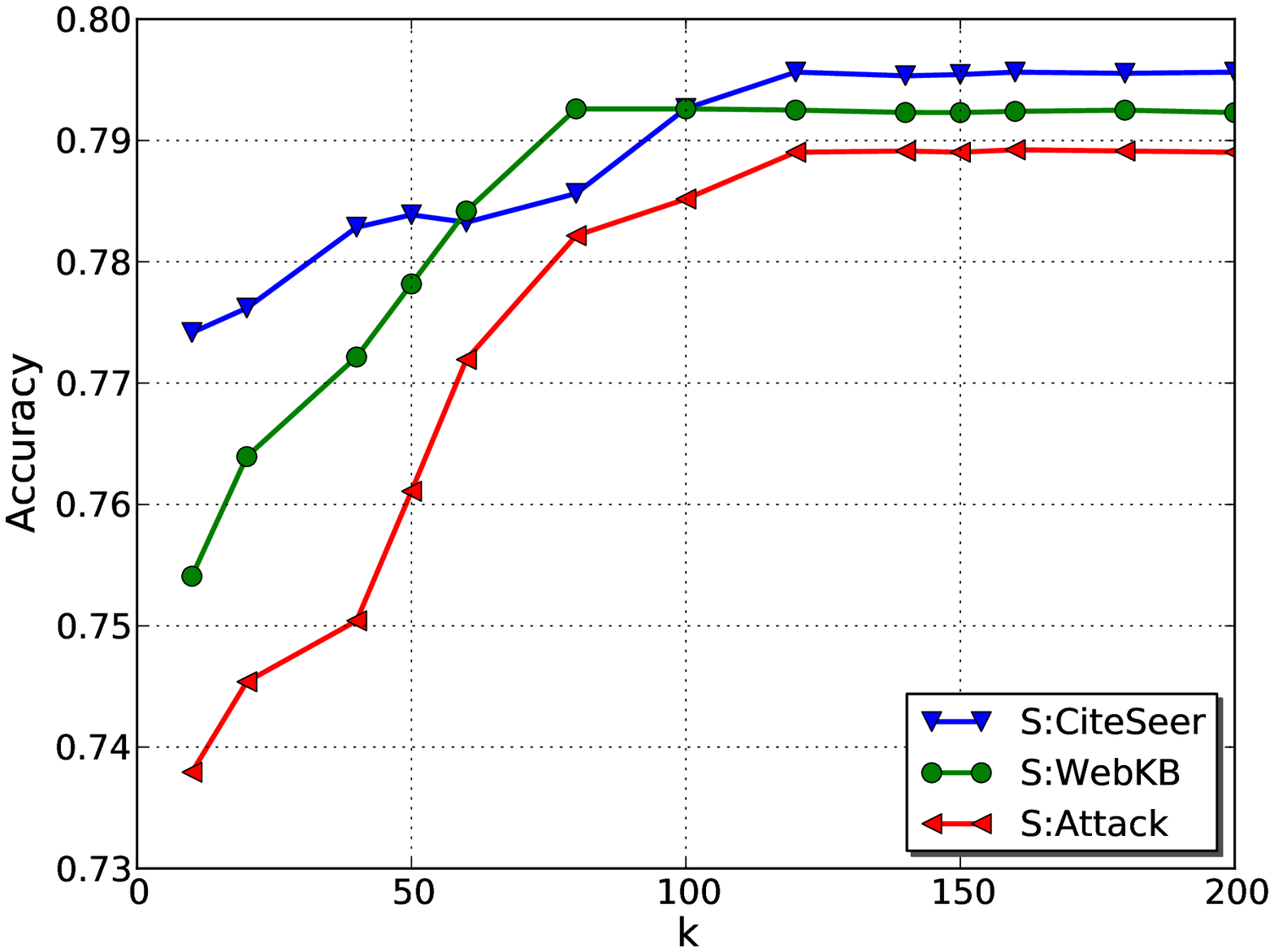}}
\subfigure[\emph{T:WebKB}]{\includegraphics[width=0.24\textwidth]{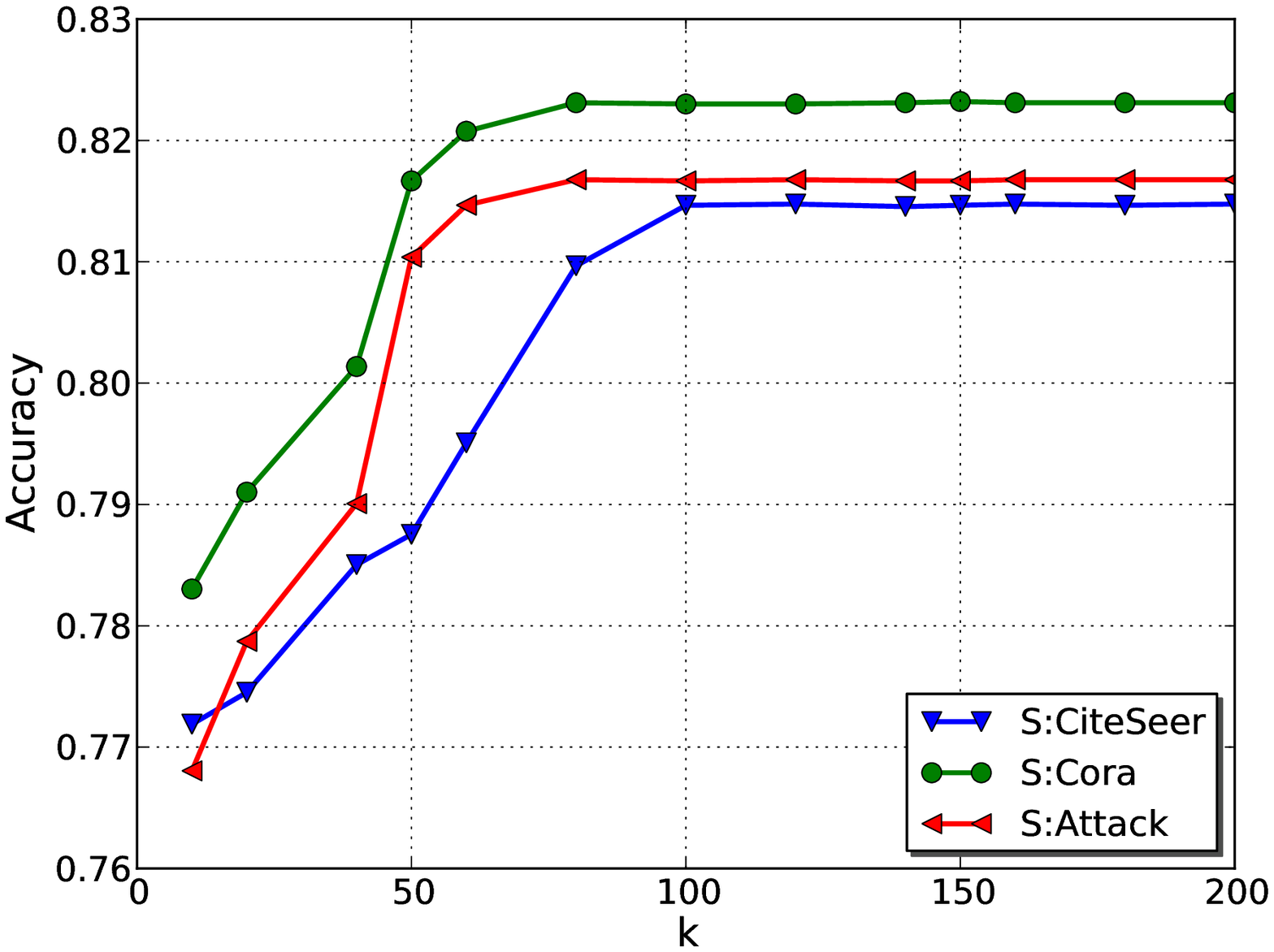}}
\subfigure[\emph{T:Attack}]{\includegraphics[width=0.24\textwidth]{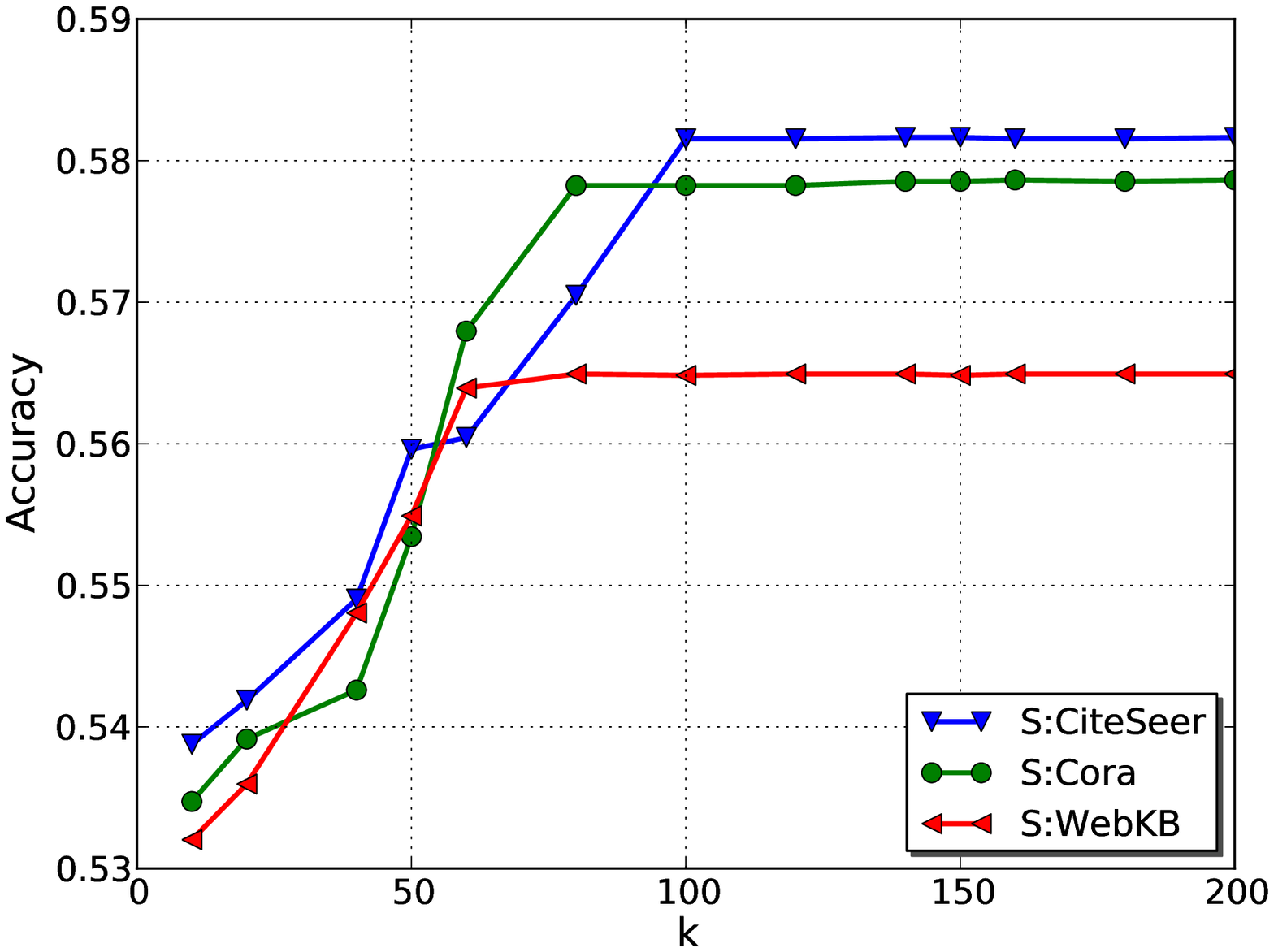}}
\caption{Classification accuracy with respect to different values of $k$. T indicates target networks and S indicates source networks.}
\label{fig-accuracy-quality}
\end{figure*}

\begin{figure*}[ht]
\centering
\subfigure[\emph{T:CiteSeer}]{\includegraphics[width=0.24\textwidth]{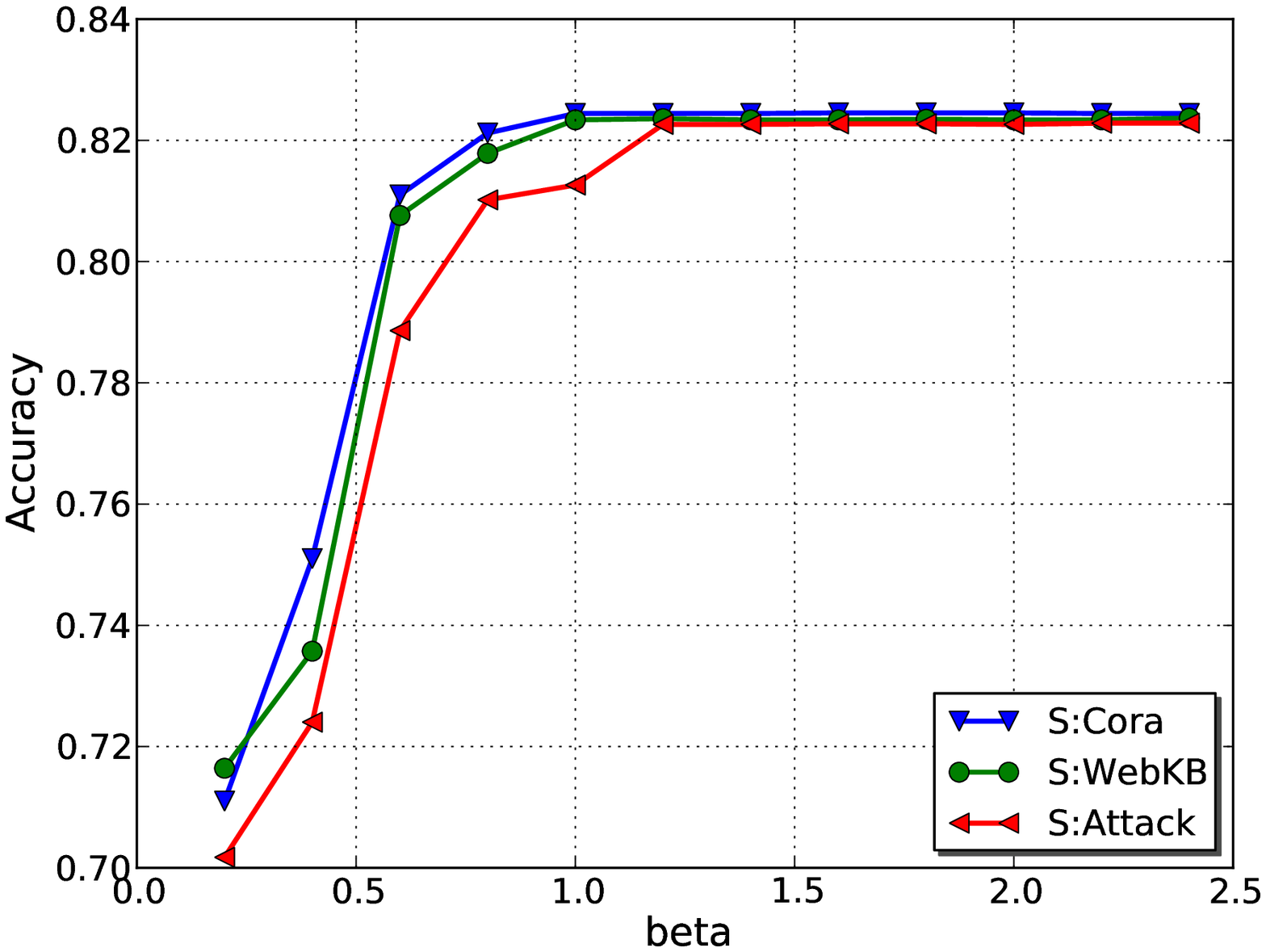}}
\subfigure[\emph{T:Cora}]{\includegraphics[width=0.24\textwidth]{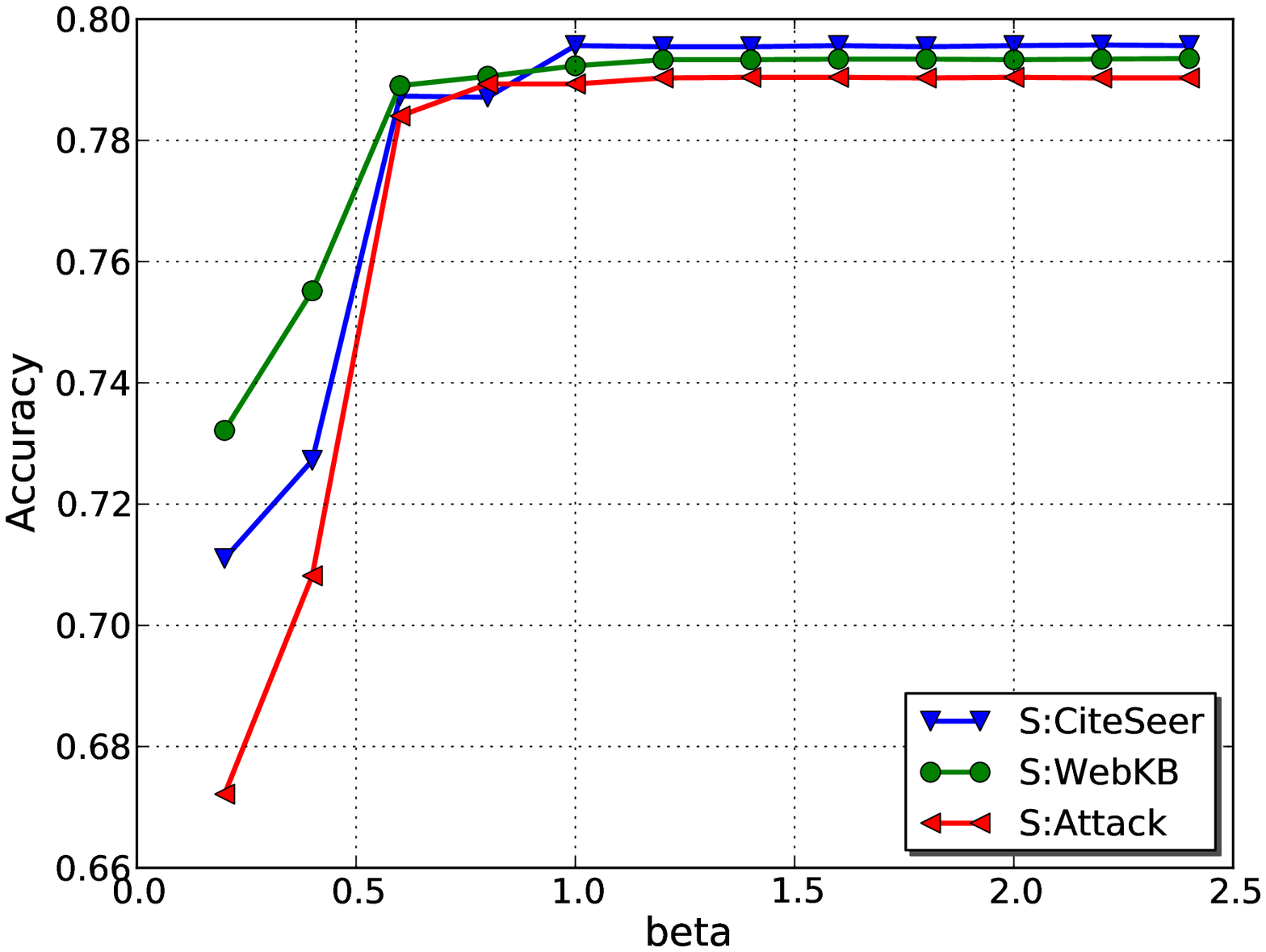}}
\subfigure[\emph{T:WebKB}]{\includegraphics[width=0.24\textwidth]{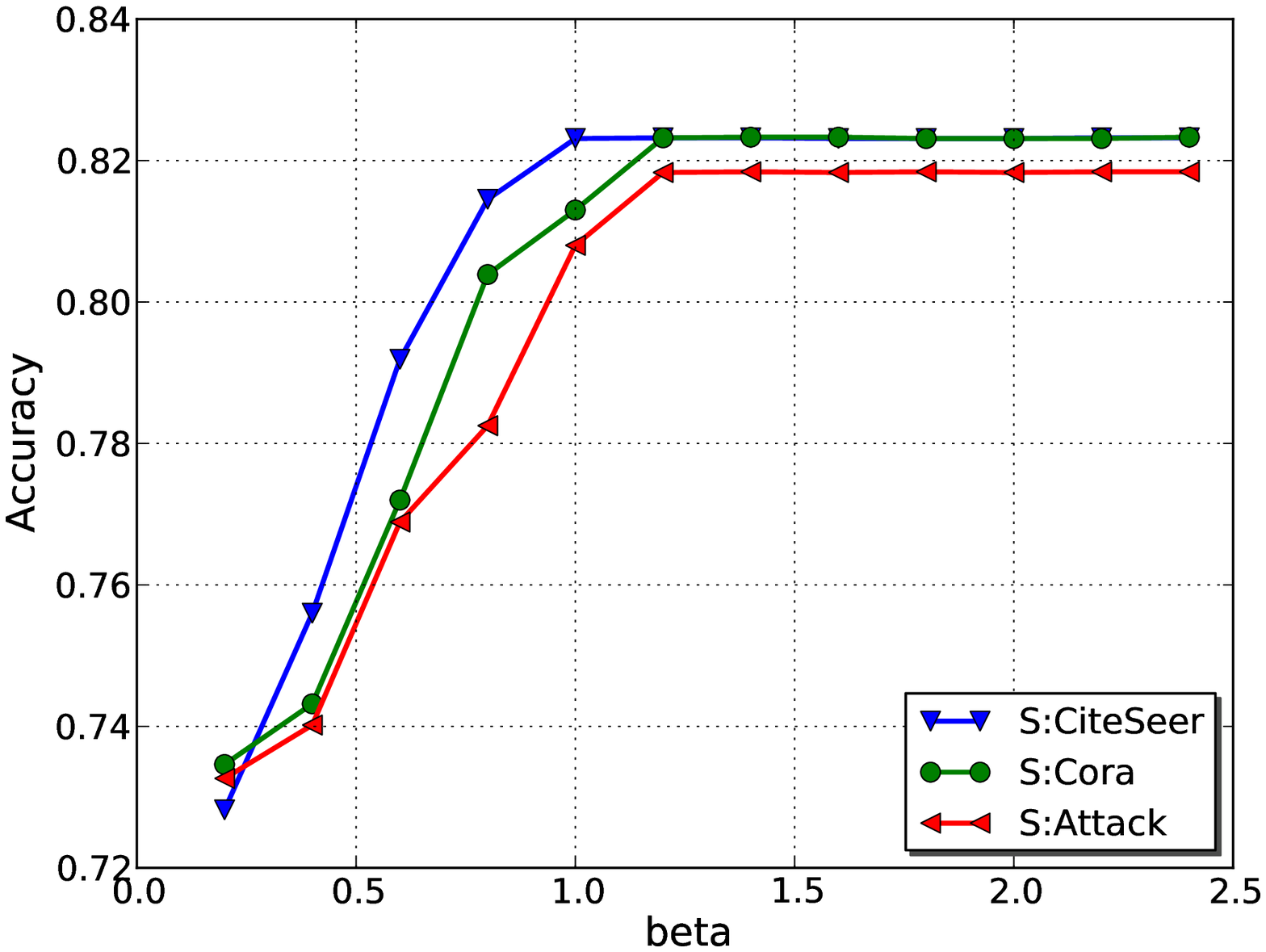}}
\subfigure[\emph{T:Attack}]{\includegraphics[width=0.24\textwidth]{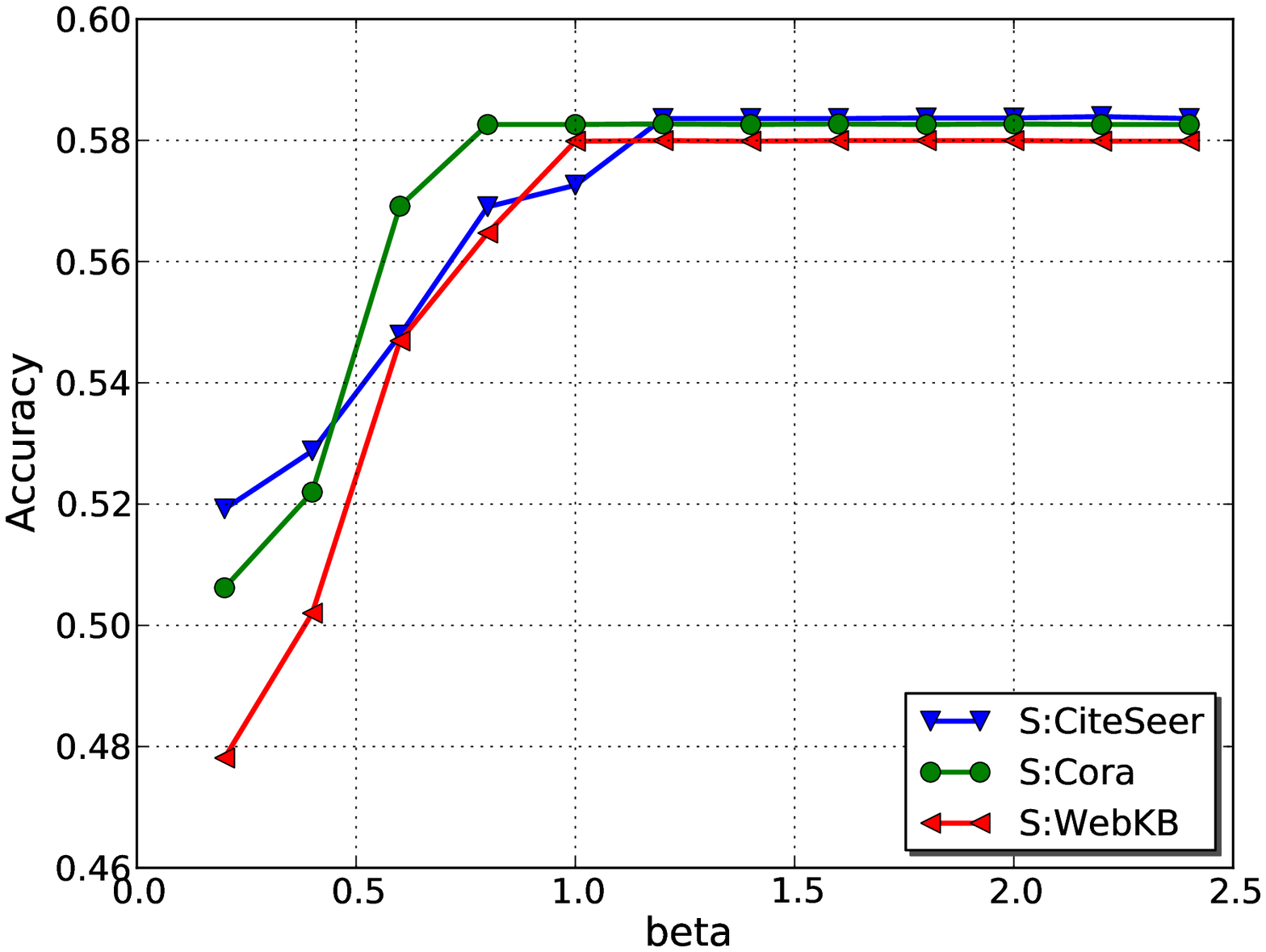}}
\caption{Accuracy comparison with different $\beta$. T indicates target networks and S indicates source networks.}
\label{fig-beta}
\end{figure*}

\subsection{Convergence of the objective function}
As the core part of our proposed TrICA algorithm, the optimization function Eq.~(\ref{eq-opt}) aims to find common latent structure features across the source and target networks. We have its derivatives to solve this optimization problem and prove that its solution can converge in Section \ref{subsec-latent-features}. Here, we also empirically validate the convergence of the objective function at different settings, where the percentage of the labeled nodes in the target network is set to be 0.5. Figure~\ref{fig-convergence} reports the values of the objective function as it converges. We can observe that the objective function can quickly converge to its optimal solution. For example, when CiteSeer is used as the target network and Cora is the source network, the value quickly decreases from $10^{4}$ to $10^{-2}$ which asserts that the objective function only takes seven iterations to converge.

\subsection{Determining the optimal value of $k$}
One important parameter of our proposed TrICA algorithm is the number of latent features $k$, when the objective function Eq.~(\ref{eq-opt}) is optimized to find the common latent structure features. Different $k$ values would lead to different feature representations used for transfer learning, and thus affect the classification accuracy on the target network. Therefore, we fix the percentage of the labeled nodes in the target network to be 0.5, and carry out experiments to test the ability of our proposed strategy to determine the optimal value of $k$.

Figure \ref{fig-quality} and Figure \ref{fig-accuracy-quality} report the quality score $\mathcal{Q}$ and classification accuracies, respectively, by varying the values of $k$. In the case that Cora is used as the target network and CiteSeer is the source network, we can see that, the maximum value of quality scores is achieved when $k$ is equal to 110, and the classification accuracy also becomes stable after $k$ reaches the value of 110. In the case that WebKB is used as the target network and CiteSeer is the source network, the maximum value of quality scores is achieved when $k$ is 110 but classification accuracy becomes stable before $k$ approaches to 110.

The results in Figures \ref{fig-quality} and \ref{fig-accuracy-quality} show that for most cases TrICA algorithm always achieves the highest accuracy when the quality score is at its local maximum value, although in some cases, the classification accuracy becomes saturated earlier before $k$ reaches its optimal values. Therefore, it still works for our requirement, because our aim is to find an optimal value of $k$ which leads to the best classification accuracy. This concludes that the local maximum value of the quality score designed in our algorithm can help decide the optimal number of latent features $k$ for achieving the best classification performance.

\subsection{Study on the impact of $\beta$}
Now we study the impact of the parameter $\beta$ on TrICA algorithm with respect to the classification accuracy. Parameter $\beta$ is a trade-off term that balances the matrix factorization and the complexity of the common feature space $A$, as defined in Eq.~(\ref{eq-opt}). For this set of experiments, we fix the percentage of the labeled nodes in the target network to be 0.5. Figure \ref{fig-beta} shows the classification accuracy by varying the $\beta$ values. We can observe that, at the beginning, as the $\beta$ value increases, TrICA achieves higher accuracies. For all the settings, when $\beta$ reaches the values between 0.5 and 1.0, the classification accuracy becomes relatively saturated. A small value of $\beta$ would relax the constraints on the values of $A$ and allow the elements in $A$ to have larger values. Consequently, this would make many values in the new features approach to become zeros in the target network, and due to the missing feature values, the node classification accuracy will deteriorate.



\section{Conclusion}
In this paper, we proposed a new algorithm to address the problem of transfer learning across different networks for node classification. We argued that for different networks the nodes' feature space and the label space can be largely (or even completely) different, and the valuable information that can be transferred is structure knowledge of the networks. Therefore, we proposed to construct a label propagation matrix to capture the influence of the structure information to the node labels in a network. Based on this idea, we formulated and solved an optimization problem to discover common latent structure features that are used for knowledge transfer. By doing so, we are able to reconstruct new structure features in the target network, which capture common structure patterns shared between networks. At the last step, an iterative classification algorithm called TrICA is proposed as the learning framework to perform collective transfer learning on the target network. Experiments and comparisons demonstrated that our proposed algorithm outperforms other baselines and the identified common latent structure features can indeed help improve the performance of collective classification for networked data.

%
%



%

%
%

\bibliographystyle{IEEEtran}
\bibliography{IEEEabrv,refs}

\end{document}